\documentclass{article}

\usepackage[preprint]{corl_2026} 

\title{\textbf{FARM}: Find Anything using\\
Relational Spatial Memory}

\usepackage{subcaption}

\usepackage{graphicx} 

\usepackage{wrapfig}   

\newcommand{\jonas}[1]{}
\newcommand{\note}[1]{}
\newcommand{\fabio}[1]{} 
\newcommand{\siming}[1]{}
\newcommand{\adam}[1]{}
\newcommand{\leo}[1]{}

\usepackage{xcolor}
\hypersetup{
    colorlinks=true,
    linkcolor=blue,
    citecolor=blue,
    urlcolor=blue
}

\usepackage{amsthm}

\usepackage{booktabs}
\usepackage{multirow}
\usepackage{makecell}
\usepackage{adjustbox}
\usepackage{amsmath,amssymb,amsfonts}
\newcommand{\parsection}[1]{\noindent\textbf{#1:}}
\usepackage{siunitx}

\usepackage{array}
\newcolumntype{L}[1]{>{\raggedright\arraybackslash}p{#1}}
\newcolumntype{C}[1]{>{\centering\arraybackslash}p{#1}}
\newcolumntype{R}[1]{>{\raggedleft\arraybackslash}p{#1}}
\usepackage[capitalize,nameinlink,noabbrev]{cleveref}
\crefname{section}{Section}{Sections}
\Crefname{section}{Section}{Sections}

\crefname{figure}{Fig.}{Figs.}
\Crefname{figure}{Fig.}{Figs.}

\crefname{table}{Table}{Tables}
\Crefname{table}{Table}{Tables}

\crefname{equation}{Eq.}{Eqs.}
\Crefname{equation}{Eq.}{Eqs.}

\crefname{question}{Question}{Questions}
\Crefname{question}{Question}{Questions}

\crefname{appendix}{Appendix}{Appendices}
\Crefname{appendix}{Appendix}{Appendices}

\newcommand{\memory}{\mathcal{M}}

\newcommand{\score}{\mathrm{score}}
\newcommand{\TopK}{\operatorname{TopK}}

\usepackage{xspace}
\newcommand{\modelnamebase}{FARM$^{\circ}$\xspace}
\newcommand{\modelname}{FARM\xspace}

\usepackage{pifont}

\usepackage{caption}
\newcommand{\xmark}{\ding{55}}

\newtheoremstyle{questionstyle}
  {0.6em}   
  {0.6em}   
  {\itshape} 
  {}        
  {\bfseries} 
  {.}       
  {0.5em}   
  {\thmname{#1}~\thmnumber{#2} \textbf{(\thmnote{#3})}} 

\theoremstyle{questionstyle}

\newtheorem{example}{Example}

\usepackage{enumitem}
\author{
\begin{tabular}{c}
Siming He$^{1}$ \quad
Leo Huang$^{1}$ \quad
Adam Lilja$^{2}$ \quad
Fabio H\"ubel$^{2}$ \quad
Jonas Frey$^{2}$
\\
Marco Pavone$^{2}$ \quad
S. Shankar Sastry$^{1}$ \quad
Jitendra Malik$^{1}$ \quad
Claire Tomlin$^{1}$
\\[0.4em]
$^{1}$UC Berkeley
\qquad
$^{2}$Stanford University
\end{tabular}
}

\usepackage{layouts}

\begin{document}
\maketitle

\begingroup
\renewcommand\thefootnote{}
\footnotetext{
\scriptsize
\noindent
\begin{minipage}{0.95\textwidth}
\raggedright
\textbf{Emails:}
\texttt{\{siminghe,klhftco\}@berkeley.edu},
\texttt{\{sastry@coe,malik@eecs,tomlin@eecs\}.berkeley.edu},
\texttt{\{adamlil,fhuebel,jonfrey\}@stanford.edu}. \textbf{Webiste:} 
\protect\href{https://goldengait.github.io/farm/}{https://goldengait.github.io/farm/}
\end{minipage}
}
\endgroup


\vspace{-3em}
\begin{figure}[htbp]
\raggedright
\includegraphics[width=\linewidth]{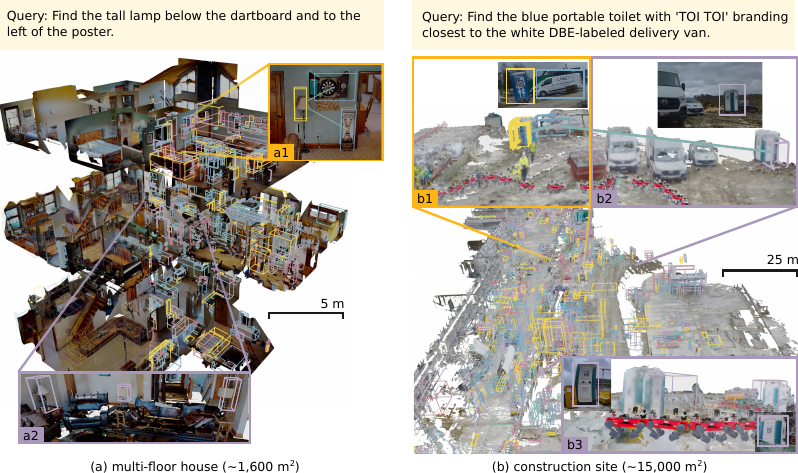}
\caption{
    \textbf{\modelname} can find objects in indoor and outdoor scenes from relational language queries.
    In (a) \protect\href{https://aihabitat.org/datasets/hm3d/00475-g7hUFVNac26/index.html}{multi-floor house}~\citep{ramakrishnan2021habitat}, it finds the lamp below the dartboard and left of the poster (a1) among 42 lamps, including visually similar distractors (a2).
    In (b) \protect\href{https://www.youtube.com/watch?v=Np2ttsGSRJA}{construction site}~\citep{frey_tuna2026grandtour}, it finds the queried portable toilet (b1), using the DBE-labeled van to disambiguate it from three same-model toilet distractors and five non-DBE-labeled vans, such as (b2) and (b3).
}
\label{fig:fig1}
\end{figure}
\vspace{-1em}

\begin{abstract}
Robots operating in homes, warehouses, and other object-rich environments need memory systems that can find specific object instances on demand.
Object-level memory alone is often insufficient: scenes contain many plausibly matching objects, and users refer to the target through relations to landmarks and surrounding objects (e.g.\ ``the tall lamp below the dartboard and to the left of the poster''), demanding a \emph{relational spatial memory} that supports retrieval through semantic, appearance, and spatial predicates over objects.
To achieve this, we present \modelname (\emph{\textbf{F}ind \textbf{A}nything using \textbf{R}elational Spatial \textbf{M}emory}), which builds, in real time at 5-10 Hz, a compact, open-vocabulary, object-level memory with geometry, visual-language descriptors, and viewpoint evidence.
At query time, \modelname uses VLMs to parse the query and score visual evidence, while grounding spatial constraints explicitly through object symbols and relational predicates.
This structured use of VLMs enables more accurate and robust retrieval than end-to-end reasoning over frame histories or scene-graph context.
In experiments on 44k language queries spanning 67 indoor and outdoor scenes, ranging from 15 to 15,000 m$^2$, \modelname improves Recall@5 and Recall@10 over prior methods by 164\% and 224\%, and a final VLM reranking stage improves Accuracy@1 by 35\%, while running in real time.
We further demonstrate closed-loop deployment on a quadrupedal robot using onboard sensors and compute.
\end{abstract}


\section{Introduction}
Consider a household robot operating in a large, object-rich house.
To be useful, it must maintain a persistent memory of the objects, landmarks, and spatial relations in its environment, and be able to answer queries about them on demand.
This requirement becomes increasingly important as robots are deployed in larger environments with many objects sharing similar categories, appearances, or functions.
Users naturally specify objects \emph{compositionally}, through their relations to landmarks, regions, and other objects (Fig.~\ref{fig:fig1}).
Recognizing each object independently is therefore insufficient.
Successful retrieval demands a \emph{relational spatial memory}: a 3D memory that treats each query not as object label lookups, but as a relational specification whose target is identified through the semantic, appearance, and spatial relations among multiple objects.
Existing systems fall short of this in different ways.

\begin{wrapfigure}[16]{r}{0.5\linewidth}
    \vspace{-1.0em}
    \centering
    \includegraphics[width=1.0\linewidth]{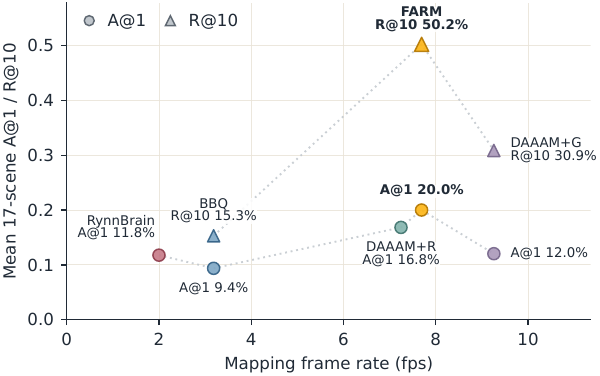}
    \caption{\textbf{Retrieval quality versus mapping throughput.}
    On ScanNet--5~\citep{dai2017scannet}, HM3D--5~\citep{ramakrishnan2021habitat}, and \modelname-Scenes, we evaluate mapping frame rate against Accuracy@1
    (circles) and Recall@10 (triangles) at visible-mask IoU
    threshold 0.1.
    }
    \label{fig:hm3d_frame_rate_vs_recall}
    \vspace{-1.0em}
\end{wrapfigure}

For instance, when it comes to memory creation, closed-vocabulary semantic SLAM and 3D scene graphs~\citep{rosinol2021kimera, hughes2022hydra, strader2024outdoor, armeni20193d, goat, sloam, activemsslam, hou2025fross} provide spatial and semantic structures but do not generalize to the open vocabularies that users naturally need.
Recent open-vocabulary scene graphs and structured memories~\citep{gu2024conceptgraphs, Schmid-RSS24-Khronos, gorlo2025DAAAM, linok2025beyond, saxena2025grapheqa, hsu2026assetcentricmetricsemanticmapsindoor, flame3d, clio, zhang2025open, zhao2026supermap} close that gap by leveraging foundation models such as CLIP~\citep{radford2021learning} and Qwen3.5~\citep{qwen35blog}; in practice, however, most still require hyperparameter tuning to operate across different scene scales, or rely on offline post-processing for memory cleaning and language grounding.
Retrieval poses a second difficulty. Earlier work~\citep{clio} relies on semantic embedding similarity (e.g., CLIP~\citep{radford2021learning}), but such embeddings typically fail to capture the relations among the multiple semantic and spatial concepts that queries invoke~\citep{gu2024conceptgraphs}.
Recent work therefore combines embeddings with foundation-model reasoning over graphs~\citep{gu2024conceptgraphs, gorlo2025DAAAM, linok2025beyond}; however, end-to-end VLM reasoning over graphs is brittle on nontrivial spatial relations, especially for smaller models that can be run onboard a robot. 
In our ablations~(\cref{tab:ablation:retrieval_rerank}), we find that end-to-end VLM reasoning can in fact \emph{hurt} performance relative to strong embedding-only retrieval.
A complementary line of work \emph{replaces} explicit memory with end-to-end VLM reasoning over raw frame histories or long video context~\citep{damo2026rynnbrain, yuan2025thinkvideosagenticlongvideo, team2024gemini, hurst2024gpt, singh2026openaigpt5card}, inheriting both the perceptual strength of modern VLMs and the limited context windows that preclude reasoning over the thousands of viewpoints accumulated in large-scale environments.
We defer a more detailed discussion of related work to Appendix~\ref{app:related_work}.

In contrast, studies by Tolman and subsequent neuroscience work suggest that animals build persistent \emph{cognitive maps}: structured spatial memories that support robust and flexible reasoning about places, objects, and relations across complex, unstructured environments~\citep{tolman1948cognitive, OKEEFE1971171, lavenex2007spatial}.
Motivated by this view, we introduce a \emph{relational spatial memory} for robots: an object-centric cognitive map that encodes semantic, geometric, and relational structure, and exposes this structure for query-time relational retrieval.
This paper addresses two coupled challenges: constructing an open-vocabulary spatial memory online at robot scale, and using that memory to retrieve object instances from free-form relational language queries.




\modelname addresses these challenges with a compact, object-centric scene graph and executable relational retrieval. During mapping, each object is represented by a single Gaussian whose updates are fully GPU-vectorized, while visual-language captioning and embedding run asynchronously off the critical path; this design uses one fixed hyperparameter set across ScanNet~\citep{dai2017scannet}, HM3D~\citep{ramakrishnan2021habitat}, and large-scale indoor--outdoor \modelname-Scenes, maintaining 5-10\,Hz online construction while improving object-grounding accuracy over prior online mapping-and-retrieval systems~\citep{linok2025beyond,gorlo2025DAAAM}. During retrieval, an LLM parses each query into a symbolic specification, explicit relational predicates score candidate bindings over memory, and a VLM resolves residual ambiguity by inspecting only a small set of viewpoints attached to the top candidates.

Across 67 indoor and outdoor scenes (ScanNet, HM3D, and our \modelname-Scenes) and 44k language queries spanning 15 to 15{,}000\,m\textsuperscript{2}, \modelname improves Accuracy@1, Recall@5, and Recall@10 by 142\%, 164\%, and 224\% over BBQ~\citep{linok2025beyond}, and Accuracy@1 by 35\% over the stronger VLM-reasoning baseline (RynnBrain~\citep{damo2026rynnbrain} or our DAAAM+RynnBrain variant). Against the costlier DAAAM~\citep{gorlo2025DAAAM}, evaluated on a 17-scene subset, it improves the same three metrics by 66\%, 65\%, and 63\% (\cref{tab:main-results}).
\Cref{fig:hm3d_frame_rate_vs_recall} summarizes the quality--throughput tradeoff, showing that \modelname achieves the highest retrieval accuracy while maintaining online mapping throughput.

In summary, our contributions are:
\begin{itemize}[noitemsep,leftmargin=*]
    \item 
    A real-time scene-graph algorithm that scales from indoor rooms to outdoor environments under one \emph{fixed} hyperparameter configuration, with no offline post-processing.
    \item 
    A retrieval framework that yields higher accuracy and better top-$K$ recall.
    \item 
    \modelname-Scenes benchmark for large-scale relational grounding, with seven curated scenes from 1{,}800 to 15{,}000\,m\textsuperscript{2}.
\end{itemize}


\section{Method}
\label{sec:asm}
We define the problem statement 
in~\cref{subsec:problem_formulation}, and describe the two key parts of \modelname: 
an online and scalable memory construction module in~\cref{subsec:memory_construction} illustrated in~\cref{fig:mapping_method}, and 
a robust relational object retrieval module leveraging constructed memory in~\cref{subsec:spatial_grounding} illustrated in~\cref{fig:retrieval_method}. More details of the system design can be found in Appendix \ref{app:detailed-method}.

\subsection{Problem Statement}
\label{subsec:problem_formulation}

\parsection{Online memory construction}
We consider a mobile robot operating in a large-scale environment. 
Up to time \(t\), the robot receives a sequence of posed RGB-D observations
\(o_{1:t}=\{o_1,\ldots,o_t\}\).
The robot constructs a persistent spatial memory online, $\memory_t = f(\memory_{t-1}, o_t),$
which summarizes the observation history for downstream retrieval.
We denote by \(\hat{\mathcal{X}}_t\) the set of retrievable object elements represented in
\(\memory_t\). 
This set is a noisy and partial approximation of the latent physical object set
\(\mathcal{X}_t\) in the environment observed up to time \(t\).

\parsection{Relational object retrieval}
Given this memory, the robot must retrieve a desired object from a natural language query
\(q \in \mathcal{Q}\), such as
\emph{\(q=\)\text{``Find the tall lamp below the dartboard and to the left of the poster.''}}
In this example, the category ``lamp'' defines an initial candidate set, while attributes
and relations such as ``tall,'' ``below the dartboard,'' and ``to the left of the poster''
disambiguate the intended object.
More generally, a query may refer to object categories, visual attributes, affordances,
spatial relations, proximity relations, or viewpoint-dependent properties.

The semantics of \(q\) induce a relational specification \(\Phi_q\).
Conceptually, \(\Phi_q\) refines an initially broad candidate set by adding descriptive
and relational constraints over the target object and any anchor objects mentioned in the query.
Let \(y \subseteq \mathcal{X}_t\) denote the query-induced set containing the desired
\emph{referent} object together with any anchor objects needed to satisfy the relational
structure of the query.
We write \(y \models \Phi_q\) when the objects in \(y\) can be assigned to the query variables
so that their categories, attributes, and mutual relations collectively satisfy the constraints
entailed by \(q\).

Given a query \(q\) and memory \(\memory_t\), the robot returns an estimated query-induced set $\hat{y} = g(q,\memory_t) \text{ and } \hat{y} \subseteq \hat{\mathcal{X}}_t$.
Retrieval succeeds when \(\hat{y}\) corresponds to the query-induced set \(y\).
One example of relational disambiguation is provided in Appendix~\ref{app:disambexample}.
This formulation leaves the implementation of \(g\) open: it may be realized by a structured
parser, embedding retriever, LLM agent, or any combination thereof.

\subsection{Memory Construction}
\label{subsec:memory_construction}

\begin{figure}[!t]
    \centering
\includegraphics[width=\linewidth]{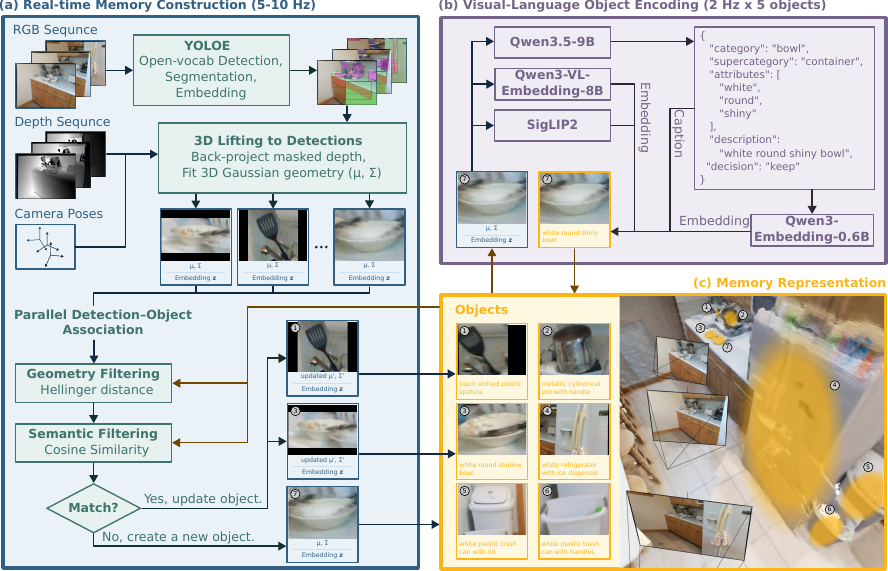}
\caption{\textbf{Online memory construction} illustrated on ScanNet~0000~\citep{dai2017scannet}.
At each timestep, \modelname runs a synchronous detect--lift--associate--fuse loop: RGB-D masks are detected, lifted with pose and depth into 3D Gaussians, matched to existing objects by geometry and appearance, and then fused or initialized as new objects.
Selected views are asynchronously captioned and embedded by VLM workers off the critical path.
The final memory stores compact object geometry, captions, retrieval embeddings, and lightweight relations for relational retrieval.}
    \label{fig:mapping_method}
\end{figure}

\parsection{Representation}
We define the relational spatial memory up to time $t$ as a set of $N_t$ entities, $\memory_t = \{\mathcal{E}_t^i\}_{i=1}^{N_t}$, where each entity $\mathcal{E}_t^i = \langle \mathcal{A}_t^i, \mathcal{R}_t^i \rangle$ stores accumulated attributes $\mathcal{A}_t^i$ and structural relations $\mathcal{R}_t^i$ to other entities.
The attribute set $\mathcal{A}_t^i$ contains (i)~a 3D Gaussian $\mathcal{N}(\mu^i,\Sigma^i)$ summarizing the entity's location and spatial extent; (ii)~a detection-time appearance feature used for cross-view association; (iii)~up to $k$ representative views, each a posed RGB crop selected for viewpoint diversity; (iv)~an open-vocabulary caption generated from those crops; and (v)~three retrieval embeddings of that evidence, consisting of a text embedding of the caption, a SigLIP2~\citep{tschannen2025siglip} image embedding, and a Qwen3-VL~\citep{qwen3vlembedding} image embedding.
The relation set $\mathcal{R}_t^i$ holds pairwise links that can be derived quickly from observations: \emph{covisibility} edges to entities that have been jointly visible in some frame, and \emph{adjacency} edges to entities that are spatially proximal under Hellinger distance.
We deliberately do \emph{not} include higher-order relations (containment, left-of, between, etc.), since the number of those relations can grow exponentially\jonas{quadratically? for fully connected graph}. Instead, we store a rich set of per-object attributes that is a compact yet sufficient summary of the scene's geometry and semantics, from which the specific relations entailed by a query can be recovered at retrieval time.

\parsection{Online construction}
As \cref{fig:mapping_method} illustrates, the robot updates the memory online as $\memory_t = f(\memory_{t-1}, o_t)$ in a loop executed once per time step given the corresponding posed RGB-D batch.
The loop maintains its frame rate by exploiting GPU parallelism on two levels.
(i)~\emph{Batched tensor operations}: frames from all cameras are stacked into a single detector forward pass, while depth back-projection, association, and fusion are implemented as vectorized CUDA operations.
(ii)~\emph{Asynchronous object captioning}: whenever the synchronous loop creates a new object or updates an object's representative views, it enqueues that object for captioning; vLLM~\citep{kwon2023efficient} captioning and embedding workers then process the selected crops and write the resulting captions and embeddings back to $\mathcal{A}_t^i$. \cref{fig:scaling-analysis} reports how mapping latency scales with scene size.

\subsection{Robust Relational Object Retrieval}
\label{subsec:spatial_grounding}


Recall from \cref{subsec:problem_formulation} that grounding a query $q$ means recovering the referent set $\hat{y} = g(q, \memory_t)$ whose elements satisfy the relational specification $\Phi_q$ induced by $q$.
Rather than providing $q$ and the full memory to a VLM and asking for an answer end-to-end, we realize $g$ in three stages: 
\emph{compile} $q$ into a small executable specification, 
\emph{score} candidate bindings of query variables to memory entities by soft unary and relational evaluators, and 
\emph{verify} the top candidates with a VLM over projected views.
Decoupling parsing, scoring, and visual verification keeps each stage efficient and lets us reason about failure modes independently. 


\begin{figure}[!t]
     \centering
    \includegraphics[width=\linewidth]{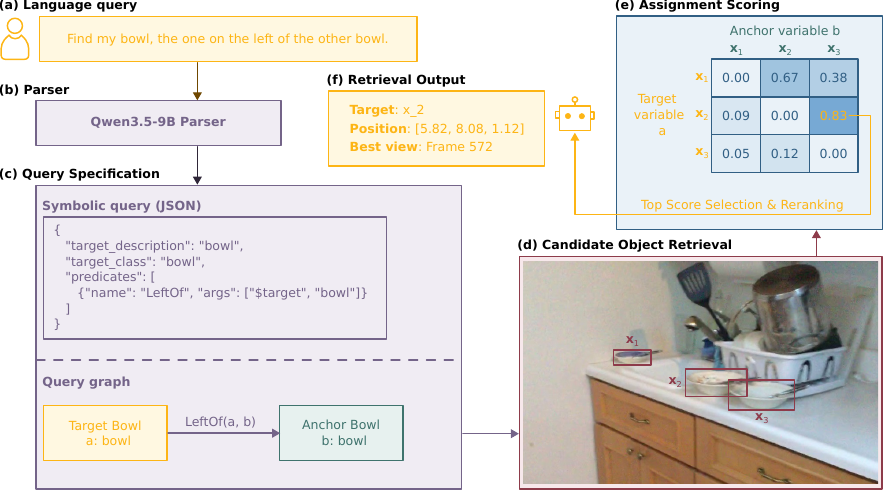}
    \caption{\textbf{Relational retrieval.}
    \modelname parses the query into a query graph of a target, anchors, and spatial predicates (a-c).
    Bindings are then scored against the memory (d) by soft predicate evaluators and reranked to return the target (e,f).
    The relation here fits in one frame; harder queries span many, where VLMs alone have no single view to reason over.}
    \label{fig:retrieval_method}
\end{figure}

\parsection{Executable query specification and compositional retrieval}
A parser compiles each query $q$ into a typed query graph
$\Pi(q)=(x_\star,{a_1,\ldots,a_m},\mathcal{S}_q,\Phi_q)$, where $x_\star$ is the target variable, $a_i$ are anchor variables, $\mathcal{S}_q$ assigns open-vocabulary descriptions to variables, and $\Phi_q$ contains spatial or semantic predicates between them. This is equivalent to a logical formula $\varphi_q(x_\star)$ with the target as the only free variable and anchors existentially quantified. Given the memory $\memory_t$, each description and predicate is evaluated softly over memory entities, producing unary scores $s_\sigma(e)\in[0,1]$ and relational scores $r_{\rho,\alpha}(e_1,\ldots,e_k)\in[0,1]$. A binding $\eta$ assigns query variables to memory entities and is scored by aggregating all induced unary and relational scores; each candidate target $e$ receives the score of its best witnessing binding,
\[
\score(e;q,\memory_t)=\max_{\eta:\eta(x_\star)=e}\mathrm{Verify}(\eta),
\qquad
\hat{y}=\TopK_{e\in\mathcal{E}_t}\score(e;q,\memory_t).
\]
Typical referring expressions form a star-shaped query graph, so conditioning on the target decomposes anchor selection and avoids exhaustive joint enumeration; details of the notation, predicate set, soft evaluators, and complexity analysis are provided in 
Appendix~\ref{app:robustretrieval}.

\parsection{Implementation}
The parser is a Qwen3.5-9B model~\citep{team2026qwen3} that emits a typed \texttt{QueryGraph} record (target description, target class, ordered predicate list).
Each unary score $s_\sigma$ is a reciprocal-rank fusion over the three retrieval embeddings stored per entity (Qwen3-text~\citep{team2026qwen3} caption text, SigLIP2~\citep{tschannen2025siglip} image, Qwen3-VL~\citep{qwen3vlembedding} image).
Each relation score $r_{\rho, \alpha}$ is a closed-form evaluator over the entity Gaussians -- Hellinger distance for \textsf{Near}, a virtual-viewer frame for \textsf{LeftOf}/\textsf{RightOf}, a vertical-offset sigmoid for \textsf{Above}/\textsf{Below}, and so on -- or a category/attribute match for the semantic predicates.
We fix each target candidate $e$ and greedily bind its anchors to their best-scoring entities, exploiting the star decomposition above.
The top-$5$ candidates are then reranked by a Qwen3.5-9B verifier that renders each candidate and its matched anchors as colored mask overlays in a previously observed view; the gap between \modelnamebase (no reranking) and \modelname in \cref{tab:main-results:quality} measures the value of this final stage.
Per-predicate evaluators and the viewer placement geometry are deferred to \cref{app:grounding-details}.

\section{Experiments}
\label{sec:experiments}

\subsection{Experimental Setup}
\label{subsec:experimental_setup}

\parsection{Datasets}
We evaluate on ReferIt3D~\citep{achlioptas2020referit_3d} built on ScanNet~\citep{dai2017scannet}, and IRef-VLA~\citep{zhang2025iref} built on HM3D~\citep{ramakrishnan2021habitat}.
For each benchmark, we select the 30 most complex scenes by object and room count, denoted ScanNet--30 and HM3D--30.
We also introduce \modelname-Scenes, a curated set of seven large-scale indoor and outdoor scenes for stress-testing scalability, covering outdoor construction, warehouse, and museum environments~\citep{frey_tuna2026grandtour, Frey-Tuna-Fu-RSS-25}, as well as a multi-floor office, campus, camping site, and factory.
For ablations and DAAAM evaluation, where GPT-5-mini cost limits full-scale runs, we use a predefined subset of 17 scenes: ScanNet--5, selected by the number of ground-truth objects; HM3D--5, selected by the number of annotated regions; and all seven \modelname-Scenes.
All methods operate online from posed RGB-D streams. Full dataset details are provided in Appendix~\ref{app:datasets}.

\parsection{Metrics and protocol}
We report Accuracy@1 (A@1), Recall@5 (R@5), Recall@10 (R@10), and mean reciprocal rank (MRR) over the top-10 candidates, as well as memory size (MiB), mapping latency (ms/frame), and query latency (ms/query).
Following RynnBrain-style evaluation~\citep{damo2026rynnbrain}, we measure success in observation space using visible-mask IoU rather than full 3D box IoU, which avoids penalizing partial online reconstructions.
For completeness, we also report 2D and 3D bounding-box IoU metrics in~\cref{tab:iou_only_additional_stats} and~\cref{tab:3d_aabb_iou_results}, where \modelname maintains the same consistent gains across evaluation protocols.
These additional metrics match prior protocols used by RynnBrain~\citep{damo2026rynnbrain} and BBQ~\citep{linok2025beyond}, respectively.
All mapping runs on a single NVIDIA RTX 5090, with VLM inference served by vLLM on an NVIDIA RTX PRO 6000. We run RynnBrain on the RTX PRO 6000 due to the VRAM requirements of its 30B model. Real-world deployments run onboard a Jetson Thor.

\parsection{Baselines}
We compare against three groups of methods: (i) object-centric scene-graph systems, including BBQ~\citep{linok2025beyond} and DAAAM~\citep{gorlo2025DAAAM}; (ii) frame-based VLM grounding, including RynnBrain~\citep{damo2026rynnbrain} and our DAAAM+RynnBrain variant, which selects frames using DAAAM to maximize object coverage; and (iii) ablations of our system, where \modelname{} denotes the full model with VLM reranking and \modelnamebase{} denotes the variant without VLM reranking.
The scene-graph baselines construct explicit maps from RGB-D streams, whereas RynnBrain grounds queries directly over subsampled frame sequences without persistent memory.
All methods use the same trajectories, queries, and evaluation pipeline. Implementation and prompt details are provided in Appendix~\ref{app:baseline_details}.

\subsection{Results}
\modelname sets a new state-of-the-art on both ScanNet~\citep{dai2017scannet} and the substantially harder HM3D~\citep{ramakrishnan2021habitat} (\cref{tab:main-results:quality}), while its rerank-free variant \modelnamebase runs about twice as fast at query time (\cref{tab:main-results:efficiency}).
%
On ScanNet--30, \modelname reaches 35.9\% A@1, a 23\% relative improvement over the strongest baseline DAAAM~\citep{gorlo2025DAAAM} (29.3\%). 
The gap widens on HM3D--30, where we improve A@1 by 30\% relative over DAAAM (7.9\% vs.\ 6.1\%).
Beyond top-1 accuracy, \modelname produces a calibrated ranking that none of the LLM-based baselines support: R@10 reaches 74.6\% on ScanNet and 26.9\% on HM3D, a $3.2\times$ and $2.1\times$ improvement over BBQ~\citep{linok2025beyond}. We additionally report performance with IoU threshold 0.25 and 0.5 in~\cref{tab:iou_thresholds_combined}. 

\cref{tab:main-results:efficiency} further shows that spatial memory construction runs at $\sim$8\,Hz, comparable to DAAAM and roughly $2\times$ faster than BBQ. 
RynnBrain~\citep{damo2026rynnbrain} reports lower per-frame latency, but on a more powerful GPU, and does not persist a scene graph as its frames are consumed transiently at query time. 
The resulting representation is compact: $\sim$23 MiB on ScanNet, smaller than every baseline except BBQ, and $\sim$125 MiB on HM3D, proportionate to the scene-complexity growth and still within typical on-device budgets. 
Query latency without reranking (1.7-2.1\,s) is competitive with DAAAM and RynnBrain and substantially faster than BBQ.

\cref{fig:scaling-analysis} confirms these properties hold at scale. 
Per-frame mapping latency remains stable as the traversed trajectory lengthens (\cref{fig:mapping-latency-scaling}), and grounding accuracy degrades gracefully as scene area grows (\cref{fig:accuracy-scene-area}), supporting use on extended on-device deployments.

\begin{table*}[t]
\centering
\caption{\modelname outperforms the baselines on all quality measures while remaining efficient for on-device use.
}
\label{tab:main-results}
\vspace{-0.45em}

\begin{subtable}[t]{\textwidth}
\centering
\tiny
\setlength{\tabcolsep}{2.7pt}
\renewcommand{\arraystretch}{0.55}
\caption{Spatial memory quality.}
\label{tab:main-results:quality}
\vspace{-0.25em}
\resizebox{\linewidth}{!}{%
\begin{tabular}{
    @{}L{1.35cm}
    *{12}{C{0.52cm}}@{}
}
\toprule
& \multicolumn{4}{c}{\textbf{ScanNet--30} (\%)}
& \multicolumn{4}{c}{\textbf{HM3D--30} (\%)}
& \multicolumn{4}{c}{\textbf{\modelname-Scenes} (\%)} \\
\cmidrule(r{2pt}){2-5} \cmidrule(lr){6-9} \cmidrule(l{2pt}){10-13}
\textbf{Method}
& A@1 & R@5 & R@10 & MRR
& A@1 & R@5 & R@10 & MRR
& A@1 & R@5 & R@10 & MRR \\
\midrule
BBQ~\citep{linok2025beyond}
& 15.4 & 22.8 & 23.5 & 18.7
&  5.3 & 10.7 & 12.8 &  7.6
&  7.4 &  9.5 &  9.6 &  8.3 \\
RynnBrain~\citep{damo2026rynnbrain}
& 28.0 & --- & --- & ---
&  4.6 & --- & --- & ---
&  2.7 & --- & --- & --- \\
DAAAM+R~\citep{gorlo2025DAAAM}
& 29.3 & --- & --- & ---
&  6.1 & --- & --- & ---
& 15.1 & --- & --- & --- \\
\modelnamebase
& 33.0 & \textbf{63.5} & \textbf{74.6} & 46.0
&  6.7 & \textbf{18.2} & \textbf{26.9} & 11.8
& 16.3 & \textbf{37.5} & \textbf{47.3} & 25.2 \\
\modelname
& \textbf{35.9} & \textbf{63.5} & \textbf{74.6} & \textbf{47.9}
&  \textbf{7.9} & \textbf{18.2} & \textbf{26.9} & \textbf{12.5}
& \textbf{24.2} & \textbf{37.5} & \textbf{47.3} & \textbf{30.5} \\ 
\midrule
DAAAM+G$\dagger$~\citep{gorlo2025DAAAM}
& 26.3 & 47.8 & 59.0 & 27.1
&  4.2 & 9.4 & 12.4 & 6.7
& 5.6 & 15.2 & 21.2 & 10.0 \\
\modelname$\dagger$
& \textbf{33.8} & \textbf{62.3} & \textbf{74.5} & \textbf{46.3}
&  \textbf{6.1} & \textbf{16.8} & \textbf{23.5} & \textbf{10.4}
& \textbf{24.2} & \textbf{37.5} & \textbf{47.3} & \textbf{30.5} \\ 

\bottomrule
\end{tabular}%
}
\end{subtable}

\vspace{0.35em}

\begin{subtable}[t]{\textwidth}
\centering
\tiny
\setlength{\tabcolsep}{3pt}
\renewcommand{\arraystretch}{0.62}
\caption{Spatial memory efficiency.}
\label{tab:main-results:efficiency}
\vspace{-0.25em}
\resizebox{\linewidth}{!}{%
\begin{tabular}{
    @{}lccc ccc ccc@{}
}
\toprule
\textbf{Method}
& \multicolumn{3}{c}{\textbf{ScanNet--30}}
& \multicolumn{3}{c}{\textbf{HM3D--30}}
& \multicolumn{3}{c}{\textbf{\modelname-Scenes}} \\
\cmidrule(r{4pt}){2-4} \cmidrule(lr){5-7} \cmidrule(l{4pt}){8-10}
& MiB & Map(ms/fr) & Query(s/q)
& MiB & Map(ms/fr) & Query(s/q)
& MiB & Map(ms/fr) & Query(s/q) \\
\midrule
BBQ~\citep{linok2025beyond}
& \textbf{10.1} & 235$\pm$106 & 3.6$\pm$0.8
& \textbf{31.7} & 314$\pm$117 & 3.7$\pm$0.9
& 31.4 & 193$\pm$134 & 4.4$\pm$2.7 \\
RynnBrain~\citep{damo2026rynnbrain}
& 67.8 & \textbf{24$\pm$12} & 2.2$\pm$0.2
& 77.0 & \textbf{26$\pm$16} & \textbf{2.1$\pm$0.2}
& 77.0 & 23$\pm$17 & 1.7$\pm$0.3 \\
DAAAM+R~\citep{gorlo2025DAAAM}
& 29.9 & \textbf{112$\pm$21} & \textbf{1.1$\pm$0.5}
& 36.2 & \textbf{138$\pm$32} & \textbf{2.1$\pm$0.6}
& \textbf{30.6} & 326$\pm$129 & \textbf{1.5$\pm$0.2} \\
\modelnamebase
& 22.1 & \textbf{121$\pm$56} & 1.7$\pm$0.3
& 120.1 & \textbf{130$\pm$57} & \textbf{2.1$\pm$0.3}
& 65.2 & \textbf{59$\pm$42} & 4.2$\pm$1.1 \\
\modelname
& 23.2 & \textbf{121$\pm$56} & 2.6$\pm$0.9
& 124.9 & \textbf{130$\pm$57} & 4.1$\pm$0.8
& 65.2 & \textbf{59$\pm$42} & 4.2$\pm$1.1 \\
\bottomrule
\end{tabular}%
}
\end{subtable}

\vspace{2pt}
\begin{minipage}{0.98\textwidth}
\footnotesize
\modelname-Scenes quality numbers use the same visible-mask IoU threshold 0.1 protocol as ScanNet~\citep{dai2017scannet}/HM3D~\citep{ramakrishnan2021habitat}. RynnBrain~\citep{damo2026rynnbrain} and our variance of DAAAM+RynnBrain are single-answer methods, so R@5/R@10/MRR are omitted.
DAAAM~\citep{gorlo2025DAAAM} (DAAAM+G) query time is $66.43 \pm 23.74\,\mathrm{s}$ per query.
\textsuperscript{$\dagger$} Results are on the same 17-scene subset as ablations~\cref{subsec:ablation}, given the token cost of GPT-5-mini~\citep{singh2026openaigpt5card}.
\end{minipage}
\vspace{-0.5em}
\end{table*}

\begin{table*}[t]
\centering
\caption{
Representation ablation evaluates whether the constructed memory contains the correct objects with useful descriptions; the retrieval ablation evaluates how different query mechanisms use the same memory. 
Methods and components: BBQ~\citep{linok2025beyond}, DAAAM~\citep{gorlo2025DAAAM}; merge backends DINO~\citep{simeoni2025dinov3}, Hydra+DAM~\citep{hughes2022hydra,lian2025describe}, YOLOE~\citep{wang2025yoloe}; retrieval embeddings Qwen3-T~\citep{team2026qwen3}, SigLIP2~\citep{tschannen2025siglip}, Qwen3-VL~\citep{qwen3vlembedding}, T5~\citep{2020t5}.
}
\label{tab:ablation:full}
\setlength{\tabcolsep}{1.0pt}
\renewcommand{\arraystretch}{0.75}
\scriptsize

\begin{subtable}[t]{0.48\linewidth}
\centering
\caption{Scene representation.}
\label{tab:ablation:representation}
\begin{tabular}{@{}lllccc@{}}
\toprule
\multirow{2}{*}{\textbf{Method}}
& \multirow{2}{*}{\textbf{Merge}}
& \multirow{2}{*}{\textbf{Retr. emb.}}
& \multicolumn{3}{c}{\textbf{A@1 / R@10}} \\
\cmidrule(lr){4-6}
& & & \textbf{ScanNet--5} & \textbf{HM3D--5} & \textbf{FARM} \\
\midrule

BBQ
& DINO
& Qwen3-T
& 5.4 / 17.2
& 2.3 / 10.3
& 5.9 / 15.3 \\

DAAAM
& Hydra+DAM
& Qwen3-T
& 18.6 / 51.0
& 2.6 / 12.2
& 5.0 / 17.43 \\

\modelname
& DINO
& Qwen3-T
& \textbf{23.0} / \textbf{58.8}
& \textbf{3.6} / \textbf{16.0}
& \textbf{9.8} / \textbf{28.8} \\

\midrule

\modelname
& DINO
& SigLIP2
& 7.4 / 30.9
& 0.3 / 1.7
& 0.9 / 3.5 \\

\modelname
& DINO
& Qwen3-VL
& 14.7 / 41.2
& 1.7 / 12.4
& 2.7 / 12.2 \\

\modelname
& DINO
& T5
& 22.1 / \textbf{59.8}
& 3.3 / 15.6
& \textbf{10.6} / 28.6 \\

\modelname
& DINO
& Qwen3-T
& 23.0 / 58.8
& 3.6 / \textbf{16.0}
& 9.8 / \textbf{28.8} \\

\modelname
& DINO
& multi
& \textbf{23.5} / 55.4
& \textbf{4.1} / 14.8
& 8.7 / 27.4 \\

\midrule

\modelname
& YOLOE
& multi
& 17.6 / 54.4
& \textbf{4.1} / \textbf{16.1}
& \textbf{9.0} / \textbf{29.0} \\

\modelname
& DINO
& multi
& \textbf{23.5} / 55.4
& \textbf{4.1} / 14.8
& 8.7 / 27.4 \\

\bottomrule
\end{tabular}
\end{subtable}
\hspace{0.6em}
\begin{subtable}[t]{0.48\linewidth}
\centering
\caption{Retrieval and reranking.}
\label{tab:ablation:retrieval_rerank}
\begin{tabular}{@{}llccc@{}}
\toprule
 \multirow{2}{*}{\textbf{Retrieval}}
& \multirow{2}{*}{\textbf{Rerank}}
& \multicolumn{3}{c}{\textbf{A@1 / R@10}} \\
\cmidrule(lr){3-5}
 & & \textbf{ScanNet--5} & \textbf{HM3D--5} & \textbf{FARM} \\
\midrule

\multicolumn{5}{l}{\emph{Retrieval mechanism ablation without reranking.}} \\

Pure emb.
& None
& 38.7 / 73.6
& 5.9 / 22.1
& 10.7 / 32.4 \\

BBQ-hybrid
& None
& 14.5 / 58.9
& 5.3 / 22.5
& 9.2 / 25.7 \\

BBQ-hybrid multi
& None
& 31.5 / 71.7
& 5.8 / 22.4
& 10.1 / 27.6 \\

Locked
& None
& \textbf{41.6} / \textbf{76.2}
& \textbf{6.7} / \textbf{24.2}
& \textbf{10.8} / \textbf{33.0} \\

\midrule
\multicolumn{5}{l}{\emph{Reranking ablation with fixed locked retrieval.}} \\

 Locked
& None
& \textbf{41.6} / \textbf{76.2}
& 6.7 / \textbf{24.2}
& 10.8 / \textbf{33.0} \\

 Locked
& Qwen@5
& 40.6 / \textbf{76.2}
& 7.9 / \textbf{24.2}
& \textbf{15.5} / \textbf{33.0} \\

 Locked
& RynnBrain@10
& 31.6 / \textbf{76.2}
& \textbf{8.4} / \textbf{24.2}
& 15.2 / \textbf{33.0} \\

\bottomrule
\end{tabular}
\end{subtable}

\vspace{-1.5em}
\end{table*}

\subsection{Ablations}
\label{subsec:ablation}
We ablate the most important components on a subset of the scenes for both spatial memory creation and retrieval.

\parsection{Spatial memory}
\cref{tab:ablation:representation} disentangles the contributions of the underlying method, the merge embedding used for cross-view association, and the retrieval embedding. 
With the merge and retrieval embeddings held fixed (DINO~\citep{simeoni2025dinov3} + Qwen3-text~\citep{team2026qwen3}), switching from BBQ~\citep{linok2025beyond} to \modelname improves A@1 by 4.3x on ScanNet--5~\citep{dai2017scannet} (5.4 $\rightarrow$ 23.0) and 1.6x on HM3D--5~\citep{ramakrishnan2021habitat}, indicating that the \modelname-mapper accounts for most of the gap to baselines in \cref{tab:main-results}. 
With the method fixed, we investigate the choice of retrieval embedding.
Among individual retrieval embeddings (HM3D--5 A@1, \%), the pure-visual SigLIP2~\citep{tschannen2025siglip} image embedding collapses (0.3), while the caption-text encoders T5~\citep{2020t5} and Qwen3-text~\citep{team2026qwen3} reach 3.3 and 3.6. Fusing the three embeddings stored per entity (Qwen3-text caption embed, SigLIP2 image embed, Qwen3-VL~\citep{qwen3vlembedding} image embed) gives the best HM3D--5 A@1 (4.1), at a small R@10 cost (16.0 → 14.8 vs. Qwen3-text alone).
Replacing DINO with YOLOE-classifier~\citep{wang2025yoloe} features for the merge costs $\sim$6 A@1 points on ScanNet while matching on HM3D, consistent with DINO being a stronger discriminator among visually similar instances.

\parsection{Retrieval and reranking}
\cref{tab:ablation:retrieval_rerank} fixes the spatial memory and ablates query mechanisms. 
The proposed \emph{locked} retrieval, multi-embedding fusion followed by the soft-predicate evaluator of \S\ref{subsec:spatial_grounding}, beats both pure-embedding cosine similarity (38.7 → 41.6 A@1 on ScanNet--5~\citep{dai2017scannet}) and BBQ's~\citep{linok2025beyond} two-stage LLM retrieval (14.5 $\rightarrow$ 41.6).
The BBQ retrieval partially recovers with multi-embedding fusion (31.5) but does not close the gap, indicating that both the fusion and the predicate scoring contribute. 
Reranking is asymmetric across benchmarks: it leaves ScanNet--5 A@1 unchanged or worse (Qwen@5: 41.6 → 40.6; RynnBrain@10: 41.6 → 31.6)~\citep{team2026qwen3,damo2026rynnbrain} but adds 1.2-1.7 points on HM3D--5~\citep{ramakrishnan2021habitat}, and never changes R@10 since it only permutes the already-retrieved top-K. 
We therefore treat reranking as a deployment knob that is useful on hard, large-scale scenes where retrieval needs a final-stage tiebreaker, but skippable on simpler benchmarks where retrieval already ranks the answer at top-1.



\begin{figure}[!t]
   \centering
    \includegraphics[width=\linewidth]{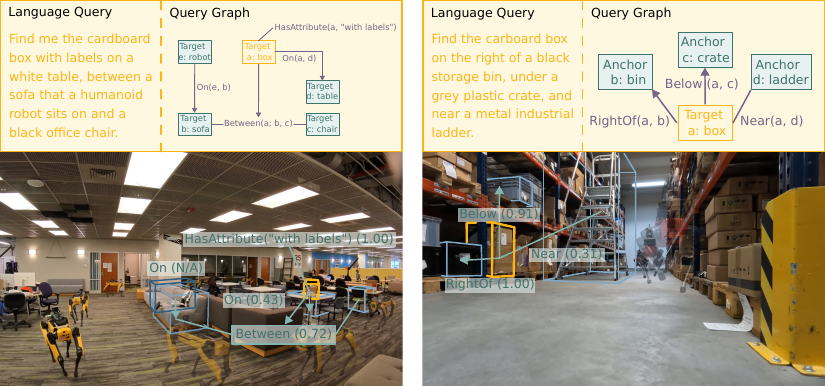}
    \caption{\textbf{Real-world relational retrieval.}
\modelname grounds natural language queries by parsing them into query graphs, binding target and anchor objects in the robot's memory, and scoring the corresponding spatial and semantic predicates.
In both indoor and warehouse scenes, the system identifies the target object among similar distractors by combining open-vocabulary attributes with relations such as \textsf{On}, \textsf{Between}, \textsf{RightOf}, \textsf{Below}, and \textsf{Near}.}
    \label{fig:realworld}
\end{figure}

\subsection{Real-world experiments}
\label{sec:realexp}

We deploy \modelname on a Boston Dynamics Spot robot equipped with onboard RGB-D sensing 
and an NVIDIA Jetson Thor compute unit. 
The full online pipeline (memory construction and relational retrieval) runs onboard without offline preprocessing or map reconstruction.
We evaluate qualitative performance on a set of natural language navigation queries in previously unseen indoor and semi-structured environments. 
The system maintains persistent object-level memory under viewpoint changes, motion blur, and partial observability, and successfully resolves relational queries involving multiple anchors and distractors. Examples are illustrated in~\cref{fig:realworld}. More details of closed-loop real-world navigation experiments are in Appendix~\ref{app:real-robot}.

\section{Conclusion}
\note{Main conclusions and future work}
We introduce \modelname, a relational spatial memory for embodied agents that jointly addresses \emph{memory construction} and \emph{memory retrieval} in large-scale open-world environments. 
By constructing this memory online, with asynchronous visual-language enrichment, and resolving each query as executable spatial predicates over the memory rather than as end-to-end reasoning over frames, \modelname overcomes the offline construction and limited scalability of prior open-vocabulary scene graphs. 
This enables real-time relational object retrieval across indoor and outdoor scenes using the same configuration across scene types. 
\modelname sets a new state-of-the-art for relational object grounding on the ScanNet~\citep{dai2017scannet} and HM3D~\citep{ramakrishnan2021habitat} referring-expression benchmarks, in both top-1 accuracy and top-$K$ recall, while mapping at roughly 8\,Hz. 
We further deploy it fully onboard a quadrupedal robot, showing that the memory can be both built and queried during operation, and release \modelname-Scenes, seven large and predominantly outdoor scenes of up to 15{,}000\,m$^2$
, to evaluate relational retrieval beyond the scale of existing benchmarks. 

\subsection{Limitations}
Our system has two main limitations.
First, the spatial predicates are manually specified and not calibrated.
Several failure cases arise from predicate scores that are too sensitive to fixed parameters; for example, \textsf{Near} can decay too quickly with distance and assign low scores to correct targets.
Similarly, we use uniform weights across semantic and spatial scores, which can cause a semantically similar distractor to dominate even when spatial constraints favor the true target.
Learning calibrated predicate functions and query-dependent score weights is an important direction for future work.

Second, current benchmarks primarily evaluate relations between the target and anchor objects, and our implementation follows this setting.
More complex queries may require reasoning over relations among anchors, such as using ``a sofa that a humanoid robot sits on'' in~\cref{fig:realworld} to disambiguate an anchor before grounding the target.
Extending both the system and benchmarks to support such compositional anchor reasoning is another promising direction.

\clearpage
\bibliography{reference}  

@article{rosinol2021kimera,
    title={Kimera: From SLAM to spatial perception with 3D dynamic scene graphs},
    author={Rosinol, Antoni and Violette, Andrew and Abate, Marcus and Hughes, Nathan and Chang, Yun and Shi, Jingnan and Gupta, Arjun and Carlone, Luca},
    journal={The International Journal of Robotics Research},
    volume={40},
    number={12-14},
    pages={1510--1546},
    year={2021},
}

@inproceedings{radford2021learning,
  title={Learning transferable visual models from natural language supervision},
  author={Radford, Alec and Kim, Jong Wook and Hallacy, Chris and Ramesh, Aditya and Goh, Gabriel and Agarwal, Sandhini and Sastry, Girish and Askell, Amanda and Mishkin, Pamela and Clark, Jack and others},
  booktitle={International conference on machine learning},
  pages={8748--8763},
  year={2021},
  organization={PmLR}
}

@misc{qwen35blog,
    title = {Qwen3.5: Accelerating Productivity with Native Multimodal Agents},
    url = {https://qwen.ai/blog?id=qwen3.5},
    author = {Qwen Team},
    month = {February},
    year = {2026}
}

@article{hughes2022hydra,
    title={Hydra: A Real-time Spatial Perception System for {3D} Scene Graph Construction and Optimization},
    fullauthor={Nathan Hughes, Yun Chang, and Luca Carlone},
    author={N. Hughes and Y. Chang and L. Carlone},
    booktitle={Robotics: Science and Systems (RSS)},
    year={2022},
}

@article{gorlo2025DAAAM,
    title={Describe Anything Anywhere At Any Moment},
    author={Nicolas Gorlo and Lukas Schmid and Luca Carlone},
    year={2025},
    eprint={2512.00565},
    archivePrefix={arXiv},
    primaryClass={cs.CV},
    url={https://arxiv.org/abs/2512.00565}
}

@inproceedings{hou2025fross,
    title={{FROSS}: {F}aster-than-{R}eal-{T}ime {O}nline 3{D} {S}emantic {S}cene {G}raph {G}eneration from {RGB-D} {I}mages},
    author={Hao-Yu Hou and Chun-Yi Lee and Motoharu Sonogashira and Yasutomo Kawanishi},
    booktitle={Proceedings of the IEEE/CVF International Conference on Computer Vision (ICCV)},
    month={October},
    year={2025},
    pages={28818-28827}
}

@article{strader2024outdoor,
    title={Indoor and Outdoor 3D Scene Graph Generation Via Language-Enabled Spatial Ontologies}, 
    author={Strader, Jared and Hughes, Nathan and Chen, William and Speranzon, Alberto and Carlone, Luca},
    journal={IEEE Robotics and Automation Letters}, 
    year={2024},
    volume={9},
    number={6},
    pages={4886-4893},
    doi={10.1109/LRA.2024.3384084}
}

@inproceedings{Schmid-RSS24-Khronos,
title     = {Khronos: A Unified Approach for Spatio-Temporal Metric-Semantic SLAM in Dynamic Environments},
author    = {Lukas Schmid and Marcus Abate and Yun Chang and Luca Carlone},
booktitle = {Proc. of Robotics: Science and Systems (RSS)},
year      = {2024},
month     = {July},
address   = {Delft, Netherlands}, 
doi       = {10.15607/RSS.2024.XX.081}
}

@article{damo2026rynnbrain,
    title={RynnBrain: Open Embodied Foundation Models},
    author={Ronghao Dang and Jiayan Guo and Bohan Hou and Sicong Leng and Kehan Li and Xin Li and Jiangpin Liu and Yunxuan Mao and Zhikai Wang and Yuqian Yuan and Minghao Zhu and Xiao Lin and Yang Bai and Qian Jiang and Yaxi Zhao and Minghua Zeng and Junlong Gao and Yuming Jiang and Jun Cen and Siteng Huang and Liuyi Wang and Wenqiao Zhang and Chengju Liu and Jianfei Yang and Shijian Lu and Deli Zhao},
    journal={arXiv preprint arXiv:2602.14979v1},
    year={2026},
    url = {https://arxiv.org/abs/2602.14979v1}
}

@misc{hsu2026assetcentricmetricsemanticmapsindoor,
      title={Asset-Centric Metric-Semantic Maps of Indoor Environments}, 
      author={Christopher D. Hsu and Pratik Chaudhari},
      year={2026},
      eprint={2510.10778},
      archivePrefix={arXiv},
      primaryClass={cs.RO},
      url={https://arxiv.org/abs/2510.10778}, 
}

@inproceedings{wang2025yoloe,
    title={YOLOE: Real-Time Seeing Anything},
    author={Wang, Ao and Liu, Lihao and Chen, Hui and Lin, Zijia and Han, Jungong and Ding, Guiguang},
    booktitle={Proceedings of the IEEE/CVF International Conference on Computer Vision (ICCV)},
    month={October},
    year={2025},
    pages={24591-24602}
}

@misc{simeoni2025dinov3,
    title={{DINOv3}},
    author={Sim{\'e}oni, Oriane and Vo, Huy V. and Seitzer, Maximilian and Baldassarre, Federico and Oquab, Maxime and Jose, Cijo and Khalidov, Vasil and Szafraniec, Marc and Yi, Seungeun and Ramamonjisoa, Micha{\"e}l and Massa, Francisco and Haziza, Daniel and Wehrstedt, Luca and Wang, Jianyuan and Darcet, Timoth{\'e}e and Moutakanni, Th{\'e}o and Sentana, Leonel and Roberts, Claire and Vedaldi, Andrea and Tolan, Jamie and Brandt, John and Couprie, Camille and Mairal, Julien and J{\'e}gou, Herv{\'e} and Labatut, Patrick and Bojanowski, Piotr},
    year={2025},
    eprint={2508.10104},
    archivePrefix={arXiv},
    primaryClass={cs.CV},
    url={https://arxiv.org/abs/2508.10104},
}

@article{tschannen2025siglip,
    title={SigLIP 2: Multilingual Vision-Language Encoders with Improved Semantic Understanding, Localization, and Dense Features},
    author={Tschannen, Michael and Gritsenko, Alexey and Wang, Xiao and Naeem, Muhammad Ferjad and Alabdulmohsin, Ibrahim and Parthasarathy, Nikhil and Evans, Talfan and Beyer, Lucas and Xia, Ye and Mustafa, Basil and H\'enaff, Olivier and Harmsen, Jeremiah and Steiner, Andreas and Zhai, Xiaohua},
    year={2025},
    journal={arXiv preprint arXiv:2502.14786}
}

@misc{team2026qwen3,
    title={Qwen3.5-Omni Technical Report}, 
    author={Qwen Team},
    year={2026},
    eprint={2604.15804},
    archivePrefix={arXiv},
    primaryClass={cs.CL},
    url={https://arxiv.org/abs/2604.15804}, 
}

@article{qwen3vlembedding,
    title={Qwen3-VL-Embedding and Qwen3-VL-Reranker: A Unified Framework for State-of-the-Art Multimodal Retrieval and Ranking},
    author={Li, Mingxin and Zhang, Yanzhao and Long, Dingkun and Chen, Keqin and Song, Sibo and Bai, Shuai and Yang, Zhibo and Xie, Pengjun and Yang, An and Liu, Dayiheng and Zhou, Jingren and Lin, Junyang},
    journal={arXiv},
    year={2026}
}

@inproceedings{achlioptas2020referit_3d,
    title={{ReferIt3D}: Neural Listeners for Fine-Grained 3D Object Identification in Real-World Scenes},
    author={Achlioptas, Panos and Abdelreheem, Ahmed and Xia, Fei and Elhoseiny, Mohamed and Guibas, Leonidas J.},
    booktitle={16th European Conference on Computer Vision (ECCV)},
    year={2020}
}

@inproceedings{zhang2025iref,
  title={IRef-VLA: A Benchmark for Interactive Referential Grounding with Imperfect Language in 3D Scenes}, 
  author={Zhang, Haochen and Zantout, Nader and Kachana, Pujith and Zhang, Ji and Wang, Wenshan},
  booktitle={2025 IEEE International Conference on Robotics and Automation (ICRA)}, 
  year={2025},
  pages={1677-1683},
  doi={10.1109/ICRA55743.2025.11127464}
}

@misc{puig2023habitat3,
    title  = {Habitat 3.0: A Co-Habitat for Humans, Avatars and Robots},
    author = {Xavi Puig and Eric Undersander and Andrew Szot and Mikael Dallaire Cote and Ruslan Partsey and Jimmy Yang and Ruta Desai and Alexander William Clegg and Michal Hlavac and Tiffany Min and Theo Gervet and Vladim{\'i}r Vondru{\v{s}} and Vincent-Pierre Berges and John Turner and Oleksandr Maksymets and Zsolt Kira and Mrinal Kalakrishnan and Jitendra Malik and Devendra Singh Chaplot and Unnat Jain and Dhruv Batra and Akshara Rai and Roozbeh Mottaghi},
    year={2023},
    archivePrefix={arXiv},
}

@inproceedings{linok2025beyond,
    title={Beyond Bare Queries: Open-Vocabulary Object Grounding with 3D Scene Graph}, 
    author={Linok, Sergey and Zemskova, Tatiana and Ladanova, Svetlana and Titkov, Roman and Yudin, Dmitry and Monastyrny, Maxim and Valenkov, Aleksei},
    booktitle={2025 IEEE International Conference on Robotics and Automation (ICRA)}, 
    year={2025},
    pages={13582-13589},
    doi={10.1109/ICRA55743.2025.11128059}
}

@misc{flame3d,
      title={Flame3D: Zero-shot Compositional Reasoning of 3D Scenes with Agentic Language Models}, 
      author={Sagar Bharadwaj and Ziyong Ma and Anurag Ghosh and Srinivasan Seshan and Anthony Rowe},
      year={2026},
      eprint={2605.09218},
      archivePrefix={arXiv},
      primaryClass={cs.CV},
      url={https://arxiv.org/abs/2605.09218}, 
}

@misc{yuan2025thinkvideosagenticlongvideo,
      title={Think With Videos For Agentic Long-Video Understanding}, 
      author={Huaying Yuan and Zheng Liu and Junjie Zhou and Hongjin Qian and Yan Shu and Nicu Sebe and Ji-Rong Wen and Zhicheng Dou},
      year={2025},
      eprint={2506.10821},
      archivePrefix={arXiv},
      primaryClass={cs.CV},
      url={https://arxiv.org/abs/2506.10821}, 
}

@article{team2024gemini,
  title={Gemini 1.5: Unlocking multimodal understanding across millions of tokens of context},
  author={Team, Gemini and Georgiev, Petko and Lei, Ving Ian and Burnell, Ryan and Bai, Libin and Gulati, Anmol and Tanzer, Garrett and Vincent, Damien and Pan, Zhufeng and Wang, Shibo and others},
  journal={arXiv preprint arXiv:2403.05530},
  year={2024}
}

@article{hurst2024gpt,
  title={Gpt-4o system card},
  author={Hurst, Aaron and Lerer, Adam and Goucher, Adam P and Perelman, Adam and Ramesh, Aditya and Clark, Aidan and Ostrow, AJ and Welihinda, Akila and Hayes, Alan and Radford, Alec and others},
  journal={arXiv preprint arXiv:2410.21276},
  year={2024}
}

@ARTICLE{clio,
  author={Maggio, Dominic and Chang, Yun and Hughes, Nathan and Trang, Matthew and Griffith, Dan and Dougherty, Carlyn and Cristofalo, Eric and Schmid, Lukas and Carlone, Luca},
  journal={IEEE Robotics and Automation Letters}, 
  title={Clio: Real-Time Task-Driven Open-Set 3D Scene Graphs}, 
  year={2024},
  volume={9},
  number={10},
  pages={8921-8928},
  doi={10.1109/LRA.2024.3451395}}

@inproceedings{zhang2025open,
  title = {{Open-Vocabulary Functional 3D Scene Graphs for Real-World Indoor Spaces}},
  author = {Zhang, Chenyangguang and Delitzas, Alexandros and Wang, Fangjinhua and Zhang, Ruida and Ji, Xiangyang and Pollefeys, Marc and Engelmann, Francis},
  booktitle = {IEEE/CVF Conference on Computer Vision and Pattern Recognition (CVPR)},
  year = {2025}
}

@article{saxena2025grapheqa,
    title={GraphEQA: Using 3D Semantic Scene Graphs for Real-time Embodied Question Answering},
    author={Saxena, Saumya and Buchanan, Blake and Paxton, Chris and Liu, Peiqi and Chen, Bingqing and Vaskevicius, Narunas and Palmieri, Luigi and Francis, Jonathan and Kroemer, Oliver},
    conference={CoRL},
    year={2025},
}

@inproceedings{armeni20193d,
    title={3D Scene Graph: A Structure for Unified Semantics, 3D Space, and Camera},
    author={Armeni, Iro and He, Zhi-Yang and Gwak, JunYoung and Zamir, Amir R and Fischer, Martin and Malik, Jitendra and Savarese, Silvio},
    booktitle={Proceedings of the IEEE International Conference on Computer Vision},
    pages={5664--5673},
    year={2019}
}

@article{
goat,
author = {Theophile Gervet  and Soumith Chintala  and Dhruv Batra  and Jitendra Malik  and Devendra Singh Chaplot },
title = {Navigating to objects in the real world},
journal = {Science Robotics},
volume = {8},
number = {79},
pages = {eadf6991},
year = {2023},
doi = {10.1126/scirobotics.adf6991},
URL = {https://www.science.org/doi/abs/10.1126/scirobotics.adf6991},
eprint = {https://www.science.org/doi/pdf/10.1126/scirobotics.adf6991},
}

@ARTICLE{sloam,
  author={Liu, Xu and Nardari, Guilherme V. and Cladera, Fernando and Tao, Yuezhan and Zhou, Alex and Donnelly, Thomas and Qu, Chao and Chen, Steven W. and Romero, Roseli A. F. and Taylor, Camillo J. and Kumar, Vijay},
  journal={IEEE Robotics and Automation Letters}, 
  title={Large-Scale Autonomous Flight With Real-Time Semantic SLAM Under Dense Forest Canopy}, 
  year={2022},
  volume={7},
  number={2},
  pages={5512-5519},
  doi={10.1109/LRA.2022.3154047}
}

@ARTICLE{activemsslam,
  author={Tao, Yuezhan and Liu, Xu and Spasojevic, Igor and Agarwal, Saurav and Kumar, Vijay},
  journal={IEEE Robotics and Automation Letters}, 
  title={3D Active Metric-Semantic SLAM}, 
  year={2024},
  volume={9},
  number={3},
  pages={2989-2996},
  doi={10.1109/LRA.2024.3363542}
}

@inproceedings{chen2020scanrefer,
    title={Scanrefer: 3d object localization in rgb-d scans using natural language},
    author={Chen, Dave Zhenyu and Chang, Angel X and Nie{\ss}ner, Matthias},
    booktitle={Computer Vision--ECCV 2020: 16th European Conference, Glasgow, UK, August 23--28, 2020, Proceedings, Part XX 16},
    pages={202--221},
    year={2020},
    organization={Springer}
}

@inproceedings{dai2017scannet,
    title={Scannet: Richly-annotated 3d reconstructions of indoor scenes},
    author={Dai, Angela and Chang, Angel X and Savva, Manolis and Halber, Maciej and Funkhouser, Thomas and Nie{\ss}ner, Matthias},
    booktitle={Proceedings of the IEEE Conference on Computer Vision and Pattern Recognition},
    pages={5828--5839},
    year={2017}
}

@article{ramakrishnan2021habitat,
  title={Habitat-matterport 3d dataset (hm3d): 1000 large-scale 3d environments for embodied ai},
  author={Ramakrishnan, Santhosh K and Gokaslan, Aaron and Wijmans, Erik and Maksymets, Oleksandr and Clegg, Alex and Turner, John and Undersander, Eric and Galuba, Wojciech and Westbury, Andrew and Chang, Angel X and others},
  journal={arXiv preprint arXiv:2109.08238},
  year={2021}
}

@misc{frey_tuna2026grandtour,
  title         = {GrandTour: A Legged Robotics Dataset in the Wild for Multi-Modal Perception and State Estimation},
  author        = {Jonas Frey and Turcan Tuna and Frank Fu and Katharine Patterson and Tianao Xu and Maurice Fallon and Cesar Cadena and Marco Hutter},
  year          = {2026},
  eprint        = {2602.18164},
  archivePrefix = {arXiv},
  primaryClass  = {cs.RO},
  url           = {https://arxiv.org/abs/2602.18164},
  note          = {\textsuperscript{*}Equal contribution (Turcan Tuna and Jonas Frey).}
}

@INPROCEEDINGS{Frey-Tuna-Fu-RSS-25,
  AUTHOR    = {Jonas Frey and Turcan Tuna and Lanke Frank Tarimo Fu and Cedric Weibel and Katharine Patterson and Benjamin Krummenacher and Matthias M{\"u}ller and Julian Nubert and Maurice Fallon and Cesar Cadena and Marco Hutter},
  TITLE     = {{Boxi: Design Decisions in the Context of Algorithmic Performance for Robotics}},
  BOOKTITLE = {Proceedings of Robotics: Science and Systems},
  YEAR      = {2025},
  ADDRESS   = {Los Angeles, United States},
  MONTH     = {July},
  NOTE      = {\textsuperscript{*}Equal contribution (Jonas Frey and Turcan Tuna and Frank Fu).}
}

@inproceedings{gu2024conceptgraphs,
    title={Conceptgraphs: Open-vocabulary 3d scene graphs for perception and planning},
    author={Gu, Qiao and Kuwajerwala, Ali and Morin, Sacha and Jatavallabhula, Krishna Murthy and Sen, Bipasha and Agarwal, Aditya and Rivera, Corban and Paul, William and Ellis, Kirsty and Chellappa, Rama and others},
    booktitle={2024 IEEE International Conference on Robotics and Automation (ICRA)},
    pages={5021--5028},
    year={2024},
    organization={IEEE}
}

@inproceedings{ekpo2026verigraph,
    title={Verigraph: Scene Graphs for Execution Verifiable Robot Planning},
    author={Ekpo, Daniel and Levy, Mara and Suri, Saksham and Huynh, Chuong and Swaminathan, Archana and Shrivastava, Abhinav},
    booktitle={Proceedings of the IEEE International Conference on Robotics and Automation (ICRA)},
    year={2026}
}

@inproceedings{sethuraman2021formal,
    title={Visual Question Answering based on Formal Logic}, 
    author={Sethuraman, Muralikrishnna G. and Payani, Ali and Fekri, Faramarz and Kerce, J. Clayton},
    booktitle={2021 20th IEEE International Conference on Machine Learning and Applications (ICMLA)}, 
    year={2021},
    pages={952-957},
    doi={10.1109/ICMLA52953.2021.00157}
}

@inproceedings{shi2019reasoning,
    title={Explainable and Explicit Visual Reasoning Over Scene Graphs},
    author={Shi, Jiaxin and Zhang, Hanwang and Li, Juanzi},
    booktitle={Proceedings of the IEEE/CVF Conference on Computer Vision and Pattern Recognition (CVPR)},
    month={June},
    year={2019}
}

@inproceedings{dai2024optimal,
    title={Optimal Scene Graph Planning with Large Language Model Guidance}, 
    author={Dai, Zhirui and Asgharivaskasi, Arash and Duong, Thai and Lin, Shusen and Tzes, Maria-Elizabeth and Pappas, George and Atanasov, Nikolay},
    booktitle={2024 IEEE International Conference on Robotics and Automation (ICRA)}, 
    year={2024},
    pages={14062-14069},
    doi={10.1109/ICRA57147.2024.10610599}
}

@inproceedings{ray2024tamp,
    title={Task and Motion Planning in Hierarchical {3D} Scene Graphs},
    author={A. Ray and C. Bradley and L. Carlone and N. Roy},
    year={2024},
    booktitle={isrr},
    nonote = {\linkToPdf{ttps://arxiv.org/pdf/2403.08094.pdf}},
    funding={DCIST,LL4FRESCO},
}

@inproceedings{zhu2021hierarchical,
    title={Hierarchical Planning for Long-Horizon Manipulation with Geometric and Symbolic Scene Graphs}, 
    author={Zhu, Yifeng and Tremblay, Jonathan and Birchfield, Stan and Zhu, Yuke},
    booktitle={2021 IEEE International Conference on Robotics and Automation (ICRA)}, 
    year={2021},
    pages={6541-6548},
    doi={10.1109/ICRA48506.2021.9561548}
}

@article{OKEEFE1971171,
title = {The hippocampus as a spatial map. Preliminary evidence from unit activity in the freely-moving rat},
journal = {Brain Research},
volume = {34},
number = {1},
pages = {171-175},
year = {1971},
issn = {0006-8993},
doi = {https://doi.org/10.1016/0006-8993(71)90358-1},
url = {https://www.sciencedirect.com/science/article/pii/0006899371903581},
author = {J. O'Keefe and J. Dostrovsky}
}

@article{tolman1948cognitive,
  author  = {Tolman, Edward C.},
  title   = {Cognitive Maps in Rats and Men},
  journal = {Psychological Review},
  year    = {1948},
  volume  = {55},
  number  = {4},
  pages   = {189--208},
  month   = jul,
  doi     = {10.1037/h0061626},
  pmid    = {18870876}
}

@article{lavenex2007spatial,
  author  = {Lavenex, Pierre and Lavenex, Pamela Banta and Amaral, David G.},
  title   = {Spatial Relational Learning Persists Following Neonatal Hippocampal Lesions in Macaque Monkeys},
  journal = {Nature Neuroscience},
  year    = {2007},
  volume  = {10},
  number  = {2},
  pages   = {234--239},
  month   = feb,
  doi     = {10.1038/nn1820},
  pmid    = {17195843}
}

@article{KASK2005165,
title = {Unifying tree decompositions for reasoning in graphical models},
journal = {Artificial Intelligence},
volume = {166},
number = {1},
pages = {165-193},
year = {2005},
issn = {0004-3702},
doi = {https://doi.org/10.1016/j.artint.2005.04.004},
url = {https://www.sciencedirect.com/science/article/pii/S0004370205000639},
author = {Kalev Kask and Rina Dechter and Javier Larrosa and Avi Dechter}
}

@inproceedings{kwon2023efficient,
  title={Efficient Memory Management for Large Language Model Serving with PagedAttention},
  author={Woosuk Kwon and Zhuohan Li and Siyuan Zhuang and Ying Sheng and Lianmin Zheng and Cody Hao Yu and Joseph E. Gonzalez and Hao Zhang and Ion Stoica},
  booktitle={Proceedings of the ACM SIGOPS 29th Symposium on Operating Systems Principles},
  year={2023}
}

@article{2020t5,
  author  = {Colin Raffel and Noam Shazeer and Adam Roberts and Katherine Lee and Sharan Narang and Michael Matena and Yanqi Zhou and Wei Li and Peter J. Liu},
  title   = {Exploring the Limits of Transfer Learning with a Unified Text-to-Text Transformer},
  journal = {Journal of Machine Learning Research},
  year    = {2020},
  volume  = {21},
  number  = {140},
  pages   = {1-67},
  url     = {http://jmlr.org/papers/v21/20-074.html}
}

@misc{singh2026openaigpt5card,
      title={OpenAI GPT-5 System Card}, 
      author={Singh, A. and others},
      year={2026},
      eprint={2601.03267},
      archivePrefix={arXiv},
      primaryClass={cs.CL},
      url={https://arxiv.org/abs/2601.03267}, 
}

@article{lian2025describe,
  title={Describe Anything: Detailed Localized Image and Video Captioning}, 
  author={Long Lian and Yifan Ding and Yunhao Ge and Sifei Liu and Hanzi Mao and Boyi Li and Marco Pavone and Ming-Yu Liu and Trevor Darrell and Adam Yala and Yin Cui},
  journal={arXiv preprint arXiv:2504.16072},
  year={2025}
}

@inproceedings{Fu_2026_funfact,
  title     = {FunFact: Building Probabilistic Functional 3D Scene Graphs via Factor-Graph Reasoning},
  author    = {Fu, Zhengyu and Zurbr{\"u}gg, Ren{\'e} and Qu, Kaixian and Pollefeys, Marc and Hutter, Marco and
               Blum, Hermann and Bauer, Zuria},
  booktitle = {Proceedings of the IEEE/CVF Conference on Computer Vision and Pattern Recognition (CVPR)},
  month     = {June},
  year      = {2026}
}

@article{yi2025viser,
  title={Viser: Imperative, web-based 3d visualization in python},
  author={Yi, Brent and Kim, Chung Min and Kerr, Justin and Wu, Gina and Feng, Rebecca and Zhang, Anthony and Kulhanek, Jonas and Choi, Hongsuk and Ma, Yi and Tancik, Matthew and Kanazawa, Angjoo},
  journal={arXiv preprint arXiv:2507.22885},
  year={2025}
}

@article{kirillov2023segany,
  title={Segment Anything},
  author={Kirillov, Alexander and Mintun, Eric and Ravi, Nikhila and Mao, Hanzi and Rolland, Chloe and Gustafson, Laura and Xiao, Tete and Whitehead, Spencer and Berg, Alexander C. and Lo, Wan-Yen and Doll{\'a}r, Piotr and Girshick, Ross},
  journal={arXiv:2304.02643},
  year={2023}
}

@misc{ray2025structuredinterfacesautomatedreasoning,
      title={Structured Interfaces for Automated Reasoning with 3D Scene Graphs}, 
      author={Aaron Ray and Jacob Arkin and Harel Biggie and Chuchu Fan and Luca Carlone and Nicholas Roy},
      year={2025},
      eprint={2510.16643},
      archivePrefix={arXiv},
      primaryClass={cs.CV},
      url={https://arxiv.org/abs/2510.16643}, 
}

@article{winograd1972understanding,
title = {Understanding natural language},
journal = {Cognitive Psychology},
volume = {3},
number = {1},
pages = {1-191},
year = {1972},
issn = {0010-0285},
doi = {https://doi.org/10.1016/0010-0285(72)90002-3},
url = {https://www.sciencedirect.com/science/article/pii/0010028572900023},
author = {Terry Winograd},
}

@inproceedings{zhao2026supermap,
  title     = {SuperMap: A Spatio-Temporal SLAM System for Visual-Language Navigation},
  author    = {Zhao, Shibo and Chen, Guofei and Zhu, Honghao and Li, Zhiheng and Yao, Changwei and Zantout, Nader and Kim, Seungchan and Wang, Wenshan and Zhang, Ji and Scherer, Sebastian},
  booktitle = {Proceedings of Robotics: Science and Systems (RSS)},
  year      = {2026}
}
\clearpage

\appendix


\section*{Supplementary Material Guide}

This supplement provides additional details to support reproducibility and to help readers navigate the evaluation protocol, implementation choices, and analyses.

\begin{itemize}[leftmargin=*,topsep=2pt,itemsep=2pt]
    \item \textbf{Related work} is discussed in~\cref{app:related_work}, with emphasis on online 3D scene graphs, open-vocabulary spatial memory, explicit retrieval over memory, and video-based VLM grounding.

    \item \textbf{Retrieval formulation and notation} are provided in~\cref{app:robustretrieval}, including the query compilation procedure, soft predicate verification, query time-complexity, and a notation reference for the logical formulation.


    \item \textbf{Implementation details} are given in~\cref{app:detailed-method}, including online memory construction, asynchronous visual-language enrichment, object association, predicate evaluation, composite scoring, and projected-view reranking.

    \item \textbf{\modelname-Scenes} are described in~\cref{app:farm-scenes}, including the scene data card, the annotation procedure for objects, and selection rationale for the large-scale indoor and outdoor environments.

    \item \textbf{Evaluation protocols and baselines} are detailed in~\cref{app:datasets,app:baseline_details}, including dataset construction, utterance sampling, HM3D trajectory rendering, baseline implementations, shared evaluation settings, and fairness controls.

    \item \textbf{Additional quantitative results} are reported in~\cref{app:evaluation-metrics,app:scaling-analysis,app:grounding_accuracy,app:failure-analysis}, including scaling analysis with respect to scene size, grounding accuracy under three criteria (visible-mask, 2D-box, and 3D-AABB IoU), natural-versus-synthetic language splits, and a failure analysis.
    
    \item \textbf{Real-robot experiment details} are provided in~\cref{app:real-robot}, including the Spot robot setup, onboard RGB-D sensing and Jetson Thor compute, deployment pipeline, representative relational queries, and qualitative analysis.
\end{itemize}

\section{Related Work}
\label{app:related_work}

Robot navigation from natural-language instructions has received growing attention as robots are deployed in larger, more complex, and more interactive environments.
A central capability in this setting is referential object grounding: given an instruction, the robot must identify the intended object instance rather than merely recognize an object category.
Several benchmarks have been introduced to evaluate this capability~\citep{achlioptas2020referit_3d,chen2020scanrefer,zhang2025iref}.
ReferIt3D~\citep{achlioptas2020referit_3d} (NR3D and SR3D) and ScanRefer~\citep{chen2020scanrefer}, built on ScanNet~\citep{dai2017scannet}, emphasize fine-grained referring expressions and spatial disambiguation among visually similar objects.
IRef-VLA~\citep{zhang2025iref} extends 3D referential grounding toward larger scenes such as HM3D~\citep{ramakrishnan2021habitat} and includes imperfect or ambiguous referring expressions.
We evaluate on both ReferIt3D and IRef-VLA.
Since these benchmarks primarily focus on everyday indoor environments, we further curate and annotate \modelname{}-Scenes: seven large-scale indoor and outdoor scenes spanning a museum, warehouse, multi-floor office building, campus, construction site, factory, and camping site.
These scenes range from 1,800 m$^2$ to 15,000 m$^2$, compared with roughly 15 m$^2$ to 1,600 m$^2$ for the ScanNet and HM3D scenes used in our evaluation.
Many prior grounding methods assume access to a pre-built point cloud or complete scene mesh.
Such assumptions are valuable for studying language grounding in reconstructed scenes, but they differ from our setting, where a robot must build and update memory online from streaming RGB-D observations and later query  that memory for object retrieval.
We therefore focus our discussion on two aspects most closely related to our work: online memory construction and retrieval over that memory.

\parsection{Explicit memory: 3D scene graphs}
3D scene graphs and metric-semantic SLAM provide a natural substrate for connecting language, semantics, and geometry in robot memory~\citep{rosinol2021kimera,hughes2022hydra,strader2024outdoor,armeni20193d,goat,sloam,activemsslam,hou2025fross}.
Armeni et al.~\citep{armeni20193d} introduced a hierarchical building/room/object/camera representation for grounding semantics in 3D, although the graph was constructed offline.
Kimera and Hydra~\citep{rosinol2021kimera,hughes2022hydra} demonstrated real-time metric-semantic SLAM and online 3D scene-graph construction on robots.
FROSS~\citep{hou2025fross} further improves efficiency by representing objects with 3D Gaussians instead of relying on dense point-cloud processing.
These systems establish important foundations for online semantic spatial memory.
At the same time, many of them rely on object classes or relation vocabularies fixed at mapping time, which limits retrieval queries that use novel object names, fine-grained attributes, or free-form referring expressions that arise after mapping.

Recent open-vocabulary mapping systems address this limitation by attaching foundation-model features or language annotations to 3D object maps.
ConceptGraphs~\citep{gu2024conceptgraphs} fuses class-agnostic 2D mask proposals into 3D object nodes, augments them with CLIP features and VLM-generated captions, and infers open-vocabulary relations between selected object pairs using an LLM.
Similarly, BBQ~\citep{linok2025beyond} builds an open-vocabulary object-centric map, with an emphasis on best-view selection for object captioning, metric graph edges, and LLM-based deductive reasoning for relational object retrieval.
FunFact~\citep{Fu_2026_funfact} builds an object- and part-centric 3D map from posed RGB-D images and then infers probabilistic functional relations through factor-graph reasoning.
These methods are closely related to our goal of open-vocabulary object grounding, but differ in when semantic and relational structure is attached to the map.
ConceptGraphs, BBQ, and FunFact construct or enrich semantic, relational, or functional graph structure in stages after the underlying 3D object representation is formed.
This map-then-enrich-then-query pipeline is powerful, but can introduce an additional offline or post-processing stage before the memory is ready for downstream retrieval.
In contrast, \modelname updates open-vocabulary object memory synchronously with online perception, so the robot can query the memory directly anytime without a separate graph-enrichment phase.
Task-dependent relations are instead instantiated at query time.
Importantly, this query-time step is not an offline relation-building pass:
the memory already stores compact object-level semantic, visual, and geometric evidence, and retrieval evaluates only the small set of predicates required by the current language instruction.
This avoids committing during mapping to a fixed set of pairwise relations, functional affordances, or task-specific predicates that may be irrelevant to future queries, making the design better suited to online robot applications where memory must be continuously updated and immediately usable.

Clio, Khronos, and DAAAM~\citep{clio,Schmid-RSS24-Khronos,gorlo2025DAAAM} move further toward online open-set scene understanding.
Clio builds compact task-conditioned open-set 3D scene graphs online, Khronos maintains real-time spatio-temporal metric-semantic maps, and DAAAM targets real-time 4D scene graphs with detailed open-vocabulary annotations.
These systems are complementary to ours and demonstrate the importance of online open-vocabulary spatial memory.
Our work focuses on a different retrieval-centric design point: arbitrary fine-grained referring-expression retrieval over large object memories.
In particular, Clio compresses the map with respect to a task distribution, Khronos emphasizes temporal metric-semantic SLAM, and DAAAM inherits scale-sensitive parameters from Hydra-style mapping, such as voxel resolution and truncation distance, which may require adjustment across scenes with substantially different spatial scales.
In contrast, \modelname{} uses a compact object-level memory whose construction does not depend on a global volumetric resolution, while still retaining fine-grained open-vocabulary descriptions for retrieval.

\parsection{Object retrieval with explicit memory}
Given an explicit memory, retrieval introduces an additional challenge: the system must combine object semantics, attributes, and spatial relations in the query.
Embedding-only retrieval methods, such as CLIP-based matching~\citep{radford2021learning,clio}, are effective for many semantic queries but often underrepresent the compositional structure of relational language, as also observed by ConceptGraphs~\citep{gu2024conceptgraphs}.
A common response is to combine embedding retrieval with foundation-model reasoning over graph context~\citep{gu2024conceptgraphs,gorlo2025DAAAM,linok2025beyond,saxena2025grapheqa}.
This is a flexible and powerful paradigm, especially when large models and rich graph annotations are available.
We therefore include the strong DAAAM retrieval baseline that uses GPT-5-mini as the reasoning model over memory, in addition to smaller VLM variants from BBQ~\citep{linok2025beyond}.
Our results suggest that, for fine-grained spatial disambiguation, increasing the reasoning model size alone does not fully address the difficulty of end-to-end inference over scene-graph context.
As shown in~\cref{tab:ablation:retrieval_rerank}, VLM re-ranking can even reduce performance relative to a strong embedding-only baseline in some settings.
We therefore use VLMs in a more structured role: the VLM parses the language instruction into a symbolic relational specification, and the resulting predicates are evaluated explicitly over \modelname memory.
This preserves the linguistic flexibility of VLMs while ensuring spatial reasoning is verifiable and less dependent on end-to-end graph-context inference.

\paragraph{Symbolic and modular reasoning over scene graphs.}
The idea of grounding language by executing symbolic operations over an explicit world model dates back at least to SHRDLU~\citep{winograd1972understanding}, which answered questions and executed commands about objects and spatial
relations in a blocks world.
Modern executable reasoning over scene graphs follow the same broad principle in richer visual domains.
XNMs~\citep{shi2019reasoning} decompose 2D visual question answering into neural modules operating over scene-graph nodes and edges, while Sethuraman et al.~\citep{sethuraman2021formal} translate 2D scene graphs and questions into Prolog facts and clauses for satisfiability checking.
These works show that, when the world representation is accurate, query-time symbolic reasoning can solve compositional visual queries.
We adopt the same principle in 3D, but in a setting where the graph entities are noisy, metric, and constructed online from robot observations.
Accordingly, our predicates are implemented as soft truth functions over object memory and composed with fuzzy operators, while VLM verification is retained for cases where geometric predicates alone are ambiguous.

\paragraph{Scene graphs as a substrate for planning and verification.}
Another line of work uses scene graphs as inputs to downstream planners or verifiers.
VeriGraph~\citep{ekpo2026verigraph} verifies LLM-generated action sequences against tabletop scene graphs; Ray et al.~\citep{ray2024tamp, ray2025structuredinterfacesautomatedreasoning} formulate task-and-motion planning over Hydra 3D scene graphs in PDDLStream; Dai et al.~\citep{dai2024optimal} compile language missions into LTL over hierarchical scene graphs; and Zhu et al.~\citep{zhu2021hierarchical} combine symbolic and geometric scene graphs for long-horizon manipulation.
These methods study how an available scene graph can support planning, verification, or temporal reasoning.
Our contribution is upstream and complementary: we construct and query a persistent open-vocabulary object memory in real time.
For this reason, our hardware experiments use a simple object-seeking controller, while integration with richer task-and-motion or temporal-logic planners is an orthogonal direction for future work.

\paragraph{Implicit memory: video understanding through VLMs.}
A complementary approach is to replace explicit object memory with end-to-end VLM reasoning over raw frame histories or long video context~\citep{damo2026rynnbrain,yuan2025thinkvideosagenticlongvideo,team2024gemini,hurst2024gpt,singh2026openaigpt5card}.
These models inherit the strong perceptual and reasoning capabilities of modern VLMs, and recent systems increasingly support long-context video understanding and spatial grounding.
For example, RynnBrain~\citep{damo2026rynnbrain} introduces a 2B/8B/30B-MoE family built on Qwen3-VL~\citep{team2026qwen3} and post-trained on egocentric video with spatio-temporal grounding primitives such as points, bounding boxes, trajectories, areas, and affordances.
We use its 30B model as a strong ``no explicit memory'' baseline.
The main trade-off is that context length scales with trajectory duration rather than with the number of persistent scene entities.
This becomes costly in long-term, large-scale environments where a robot may accumulate thousands of viewpoints.
We also evaluate a stronger variant that combines RynnBrain with DAAAM-style~\citep{gorlo2025DAAAM} frame selection to choose a subset of frames that maximizes object coverage.
This offline selection improves over raw long-video reasoning, but still underperforms \modelname, especially in larger scenes where compact, queryable object memory becomes increasingly important.

\section{Robust Relational Retrieval Formulation}
\label{app:robustretrieval}
\parsection{Query compilation}
\Cref{tab:logic-notation} summarizes the notation used in this formulation.
A parser compiles $q$ into a typed specification
$\Pi(q) = (x_\star, \{a_1, \ldots, a_m\}, \mathcal{S}_q, \Phi_q)$,
where $x_\star$ is the \emph{target} variable, $a_1, \ldots, a_m$ are \emph{anchor} variables, $\mathcal{S}_q$ attaches to each target and anchor variable an open-vocabulary description $\sigma$ (e.g.\ ``tall lamp'', ``white DBE delivery van''), and $\Phi_q$ lists relational predicates over those variables.
We write $s_\sigma(v)$ for the unary predicate that variable $v$ satisfies description $\sigma$, and each relation as $\rho(\mathbf{v};\alpha)$, where $\rho$ is the predicate name, $\mathbf{v}$ is the tuple of variables it relates, and $\alpha$ denotes any additional argument or parameter.
Equivalently, $q$ becomes a logical formula with one free variable,
\[
    \varphi_q(x_\star) = \exists\, a_1, \ldots, a_m \;\Big( \bigwedge_{(v, \sigma) \in \mathcal{S}_q} \!\! s_\sigma(v) \;\wedge\; \bigwedge_{\rho(\mathbf{v}; \alpha) \in \Phi_q} \!\! \rho(\mathbf{v}; \alpha) \Big),
\]
so the target is the only free variable and anchors exist only to constrain it through the predicates that mention them.
Predicate \emph{names} are selected from a fixed, but flexible set of sixteen spatial and semantic relations (\textsf{Near}, \textsf{On}, \textsf{Above}, \textsf{Below}, \textsf{NextTo}, \textsf{Between}, \textsf{Inside}, \textsf{InRegion}, \textsf{LeftOf}, \textsf{RightOf}, \textsf{InFrontOf}, \textsf{Behind}, \textsf{Closest}, \textsf{Farthest}, \textsf{HasAttribute}, \textsf{IsCategory}), while predicate \emph{arguments} -- target descriptions, attribute strings, region phrases -- remain open-vocabulary.
The parser handles spatial relations outside the closed-set by decomposing them into combinations of the sixteen base relations.
For the example in \cref{fig:fig1}, $\Pi(q)$ has $\mathcal{S}_q = \{x_\star \mapsto \text{``tall lamp''}, a_1 \mapsto \text{``dartboard''}, a_2 \mapsto \text{``poster''}\}$ and $\Phi_q = \{\textsf{Below}(x_\star, a_1), \textsf{LeftOf}(x_\star, a_2)\}$. 

\parsection{Soft verification}
The memory $\memory_t$ supplies the interpretation of the logical formula $\varphi_q$: each entity $\mathcal{E}_t^i$ exposes feature functions (location, 3D extent, embeddings, captions, views, covisibility) read from its stored attributes $\mathcal{A}_t^i$ and relations $\mathcal{R}_t^i$.
Against these, a description is scored as a soft unary match and a predicate as a soft relation $s_\sigma(e) \in [0,1] \text{ and s} r_{\rho,\alpha}(e_1,\ldots,e_k) \in [0,1],$
where $e,e_1,\ldots,e_k \in \mathcal{E}_t$, $r_{\rho,\alpha}$ is the soft score for predicate $\rho(\cdot;\alpha)$, and $k$ is the arity of $\rho$.
Thus, perception and language uncertainty enter as preferences rather than as hard filters.
A \emph{binding} $\eta: \{x_\star, a_1, \ldots, a_m\} \to \mathcal{E}_t$ assigns each variable a memory entity, and is verified by aggregating the unary and relational scores it induces,
\[
    \mathrm{Verify}(\eta) = \operatorname{Agg}\!\Big( \{ s_\sigma(\eta(v)) \}_{(v, \sigma) \in \mathcal{S}_q} \,\cup\, \{ r_{\rho, \alpha}(\eta(\mathbf{v})) \}_{\rho(\mathbf{v};\alpha) \in \Phi_q} \Big),
\]
where $\eta(\mathbf{v})$ applies the binding componentwise to the tuple $\mathbf{v}$, and $\operatorname{Agg}$ combines the semantic and relational scores into a single soft constraint-satisfaction score.
A candidate target $e$ is scored by its best witnessing binding, and retrieval returns the top-$K$:
\[
    \score(e; q, \memory_t) = \max_{\eta:\, \eta(x_\star) = e} \mathrm{Verify}(\eta), \qquad \hat{y} = \TopK_{e \in \mathcal{E}_t} \score(e; q, \memory_t).
\]
An optional action interface $\Psi_t$ then maps the chosen entity to a navigable or viewing pose that robot goes to.

\parsection{Query-time complexity}
After candidate generation, binding cost depends on the query factor graph and the candidate-set sizes.
A typical referring expression forms a \emph{star}: one target $x_\star$ connected to $m$ anchors $a_i$ by independent predicates $\rho_i(x_\star, a_i)$.
For a decomposable score, conditioning on $x_\star$ separates the anchors,
\[
    \max_{x_\star, a_1, \ldots, a_m} \sum_{i=1}^m \rho_i(x_\star, a_i) = \max_{x_\star} \sum_{i=1}^m \max_{a_i} \rho_i(x_\star, a_i),
\]
so exact binding on a star costs $O(|\Phi_q|\, K_t K_a)$, where $K_t, K_a$ bound the target and per-anchor candidate counts -- rather than the $O(K_t K_a^m)$ of exhaustive joint enumeration.
Higher-arity predicates such as $\textsf{Between}(x_\star, a_p, a_q)$ couple only the variables they mention; more generally, exact bucket elimination on a query factor graph of treewidth $b$ costs $O(|\Phi_q|\, K^{b+1})$~\citep{KASK2005165}.

\subsection{Notation Reference for the Retrieval Formulation}
\label{app:logics-notations}

This appendix uses first-order logic notation only as a compact way to specify
which memory entities should be considered valid interpretations of a query.
The notation should be read operationally as a soft constraint-satisfaction
problem over memory entities, rather than as a hard symbolic logic system.

\begin{table}[h]
\centering
\small
\begin{tabular}{p{0.24\linewidth}p{0.68\linewidth}}
\toprule
\textbf{Symbol} & \textbf{Meaning} \\
\midrule
$q$ & Natural-language referring query. \\
$\Pi(q)$ & Parsed query specification. \\
$x_\star$ & Target variable: the object to retrieve. \\
$a_1,\ldots,a_m$ & Anchor variables: contextual objects used to identify the target. \\
$\mathcal{S}_q$ & Variable-to-description map, e.g., $x_\star \mapsto$ ``tall lamp''. \\
$\Phi_q$ & Set of relational predicates, e.g., $\textsf{Below}(x_\star,a_1)$. \\
$s_\sigma(v)$ & Unary description predicate: variable $v$ matches text description $\sigma$. \\
$\rho(\mathbf{v};\alpha)$ & Relation predicate with variables $\mathbf{v}$ and optional argument $\alpha$. \\
$\varphi_q(x_\star)$ & Logical form of the query, with $x_\star$ as the only free variable. \\
$\eta$ & Binding from query variables to memory entities. \\
$r_{\rho,\alpha}$ & Soft score for relation $\rho(\cdot;\alpha)$. \\
$\mathrm{Verify}(\eta)$ & Compatibility score of a complete binding. \\
$\score(e;q,\memory_t)$ & Best verification score when entity $e$ is fixed as the target. \\
\bottomrule
\end{tabular}
\caption{Notation used in the robust relational retrieval formulation.}
\label{tab:logic-notation}
\end{table}

To read the formula for $\varphi_q(x_\star)$, interpret
$\exists a_1,\ldots,a_m$ as ``choose some anchor entities from memory,''
$\bigwedge$ as ``satisfy all listed descriptions and relations jointly,''
and $x_\star$ as the only variable whose binding is returned as the answer.
For example, in ``the lamp below the dartboard and to the left of the poster,''
the lamp is the target, while the dartboard and poster are anchors that help
disambiguate it.

Unlike classical logic, these predicates are evaluated softly. A description
match $s_\sigma(e)$ and a relation match $r_{\rho,\alpha}(e_1,\ldots,e_k)$
produce scores in $[0,1]$ rather than binary truth values. Thus,
$\mathrm{Verify}(\eta)$ measures how well one complete assignment of variables
to memory entities satisfies the query, and $\score(e;q,\memory_t)$ keeps the
best such assignment among all bindings that choose $e$ as the target.

\section{Relational Disambiguation Example}
\label{app:disambexample}
\begin{example}[Relational disambiguation]
For the query
\[
    q=\text{``Find the tall lamp below the dartboard and to the left of the poster,''}
\]
the category ``lamp'' first defines a broad candidate set of possible referents,
\[
    \mathcal{C}_0 = \{x \in \mathcal{X}_t : x \text{ is a lamp}\}.
\]
In a large scene, \(\mathcal{C}_0\) may contain many lamps. The additional
constraints ``tall,'' ``below the dartboard,'' and ``to the left of the poster''
progressively narrow this candidate set to the intended referent. The resulting
query-induced grounding set is
\[
    y =
    \{x_{\mathrm{lamp}}, x_{\mathrm{dartboard}}, x_{\mathrm{poster}}\}
    \subseteq \mathcal{X}_t,
\]
where \(x_{\mathrm{lamp}}\) is the target object and
\(x_{\mathrm{dartboard}}\), \(x_{\mathrm{poster}}\) are anchor objects. These
objects collectively satisfy the query specification,
\[
    y \models \Phi_q,
\]
because \(x_{\mathrm{lamp}}\) is a tall lamp, \(x_{\mathrm{lamp}}\) is below
\(x_{\mathrm{dartboard}}\), and \(x_{\mathrm{lamp}}\) is to the left of
\(x_{\mathrm{poster}}\). Thus, the relational query disambiguates the desired
lamp by shrinking the referent candidates while grounding the anchor objects
needed to identify it.
\end{example}

\section{Detailed Method}
\label{app:detailed-method}

\subsection{Memory Construction Details}
\label{app:memory-construction}

This appendix expands the two stages of the online memory update introduced in \cref{subsec:memory_construction}: the synchronous per-frame loop that runs on the robot's critical path, and the asynchronous visual-language enrichment that runs off it.

\paragraph{Per-frame synchronous update.}
For each incoming RGB-D batch, the pipeline executes the following seven steps, followed by a periodic pruning pass.
\begin{enumerate}
\item \emph{Segmentation.} An open-vocabulary detector~\cite{wang2025yoloe} produces per-image masks together with classification-head features used as compact per-detection embeddings. Each detection is lifted into a 3D Gaussian $(\mu_d, \Sigma_d)$ over the masked depth points, with $\Sigma_d$ stored in a 6-vector packed form. An optional dense backbone~\cite{simeoni2025dinov3} provides per-token features that are mean-pooled over the mask when richer visual descriptors are desired.
\item \emph{Filtering.} A cascade of cheap rejection tests removes masks touching the image border, masks with too few foreground pixels, masks beyond the reliable depth range, detections labeled with uninformative open-vocabulary classes (e.g.\ walls and floors), and finally near-duplicate masks (IoU). The retained set is the input to association.
\item \emph{Neighbor search.} For each surviving detection we first restrict candidates to active entities whose per-detection feature is sufficiently similar, then compute the Hellinger distance between the detection's 3D Gaussian and each candidate entity's Gaussian~\cite{hou2025fross}, retaining every candidate entity within a fixed Hellinger budget, sorted by distance. The feature/class prefilter is a single batched detection-by-entity similarity matrix, so the more expensive Gaussian comparison runs only on the resulting sparse candidate set rather than on all detection--entity pairs. The result is a sparse detection-to-entity match graph that feeds association; a detection retaining two or more entities is what later bridges previously separate entities.
\item \emph{Correspondence.} A union-find pass over the match graph with path compression resolves the case where multiple detections claim the same entity, and the case where one detection bridges two previously separate entities. 
The pass only assigns winner/loser pointers; the actual entity-level merge---in which the loser's covisibility, captions, and identity history are absorbed by the winner---is carried out during the subsequent fusion step.
\item \emph{Fusion.} Matched detections update the entity's running Gaussian by sufficient-statistics merge ($M_1 = w\mu$, $M_2 = w(\Sigma + \mu\mu^\top)$ accumulated over detections), update the entity's per-object appearance feature by detection-count weighted average. Entity-level merges flagged by the correspondence pass are also executed here, absorbing the loser's covisibility, captions, view records, and identity history into the winner. Unmatched detections initialize new entities.
\item \emph{High-quality view selection.} A new viewpoint is retained for an entity only if its viewing direction differs from all existing retained viewpoints by more than a fixed angular threshold, or if it is sufficiently closer than the closest existing viewpoint. Retained viewpoints are the inputs to the asynchronous caption stage.
\item \emph{Covisibility update.} A bitset adjacency ($N \times \lceil N/64 \rceil$ blocks) marks all pairs of entities that were jointly visible in the current batch; a parallel weighted edge dictionary accumulates their co-occurrence, restricted to spatially proximal pairs (a $k$-nearest-neighbor graph in Gaussian Hellinger distance with a soft median cutoff) and discounted by a temporal decay. This supplies the relational substrate $\mathcal{R}_t$ used during grounding.
\end{enumerate}

\paragraph{Asynchronous visual-language enrichment.}
The synchronous loop dispatches an enrichment task for any entity whose high-quality view set changed in step~6. A background worker dequeues the task, sends the selected crops to the captioning VLM~\note{\cite{team2026qwen3}}, and parses a short structured-JSON caption. It then computes the three retrieval embeddings stored per object: the caption text is embedded by Qwen3-Embedding-0.6B, while the selected crop is embedded by both a SigLIP2 image encoder~\note{\cite{tschannen2025siglip}} and Qwen3-VL-Embedding-2B. The results are written back to the entity's caption and embedding fields. Because the mapping loop is never blocked, the steady-state frame rate is set by the synchronous loop alone, not by VLM throughput; the queue may grow transiently under object saturation and is drained when the system is idle. 

\paragraph{Why a single Gaussian per entity?}
Object geometry could be stored as a point cloud, a voxel grid, or a parametric shape.
We choose a single 3D Gaussian per entity for three reasons.
First, it is \emph{resolution-free}: point and voxel representations require a global density or edge length, and the appropriate value differs by orders of magnitude between a coffee mug (centimeters) and a building fa\c{c}ade (tens of meters), so a single setting either exhausts memory outdoors or dissolves small objects indoors.
A Gaussian carries no such hyperparameter.
Second, it admits an \emph{$O(1)$ online update} in closed form, so both the per-entity footprint and the per-update cost are constant in observation count.
Third, it yields a \emph{scale-invariant matching metric}: the Hellinger distance between two Gaussians is unitless and bounded in $[0,1]$, so a single data-association threshold transfers across scenes spanning $10^1$ to $10^4$ square meters without retuning.
Together, these properties are what enable \modelname to generalize across ScanNet rooms, HM3D apartments, and outdoor construction sites under one fixed configuration (\cref{sec:experiments}).

\subsection{Retrieval Details}
\label{app:grounding-details}

This appendix expands the grounding pipeline summarized in \cref{subsec:spatial_grounding}. Each stage either narrows the candidate set or refines the score; the predicates named by the query -- not the full inventory of relations -- determine the work done at query time.

\begin{enumerate}
    \item \emph{Query parsing.}
    A Qwen3.5-9B LLM~\note{\cite{team2026qwen3}} parses the query string into a typed \texttt{QueryGraph} record holding a modifier-rich target description (e.g.\ ``red chair''), a bare target-class noun (``chair''; recovered by regex tokenization when the model is unsure), and an ordered list of predicate invocations. Predicate \emph{names} are drawn from a fixed vocabulary of sixteen spatial and semantic relations, \textsf{Near}, \textsf{On}, \textsf{Above}, \textsf{Below}, \textsf{NextTo}, \textsf{Between}, \textsf{Inside}, \textsf{InRegion}, \textsf{LeftOf}, \textsf{RightOf}, \textsf{InFrontOf}, \textsf{Behind}, \textsf{Closest}, \textsf{Farthest}, \textsf{HasAttribute}, \textsf{IsCategory}, while predicate \emph{arguments} (target and landmark descriptions, attribute strings, region labels) are open-vocabulary natural-language phrases. Since we are using LLM for parsing, our algorithm can handle open-vocabulary spatial relations by decomposing them into ones from the set of sixteen relations.
    \item \emph{Region scoping.}
    If the parsed query carries a region constraint and the scene provides region labels, the region phrase is embedded and matched against the stored labels (exact match, then cosine fallback), and entities outside the matched region are dropped from the candidate pool. This stage is skipped when the query has no region constraint or the scene has no region annotations.
    \item \emph{Candidate retrieval.}
    The candidate pool is the set of \emph{active} entities -- those flagged active in the scene state, carrying a caption and the per-object embeddings -- within the scoped region. Rather than truncating to a top-$k$, we rank the entire pool by reciprocal-rank fusion (each channel contributes $w/(\mu + \mathrm{rank})$ with offset $\mu{=}5$) over four query channels spanning three embedding spaces: the caption-text embedding (Qwen3-Embedding-0.6B) queried with the modifier-rich target description ($w{=}1.0$) and again with the raw utterance ($w{=}0.5$); the SigLIP2 image embedding queried with a short text-prompt ensemble ($w{=}0.75$); and the Qwen3-VL image embedding queried with the target description ($w{=}1.0$). The fused rank score is the entity's target similarity $s_{\mathrm{sem}}$; landmark variables referenced by the predicates are retrieved analogously.
    \item \emph{Predicate evaluation.}
    Each predicate is scored by a closed-form evaluator over the entity Gaussians, returning a soft membership in $[0,1]$: \textsf{Near} uses a Gaussian on the inter-centroid distance ($\sigma{=}0.5$\,m) and \textsf{NextTo} a wider Gaussian on the same distance ($\sigma{=}1.0$\,m); \textsf{Above}/\textsf{Below} a sigmoid on the vertical offset; \textsf{On} a horizontal-distance gate times a vertical band; \textsf{Inside}/\textsf{InRegion} a containment test; \textsf{Closest}/\textsf{Farthest} a rank within the candidate set ($\textrm{score}=1/(|\textrm{rank}-r^\star|+1)$); and the viewpoint-dependent \textsf{LeftOf}, \textsf{RightOf}, \textsf{InFrontOf}, \textsf{Behind} are evaluated in stored camera views of the candidate of the target object -- left/right from the anchor-relative horizontal offset of mask centroids on the image plane, front/behind from relative camera-frame depth. \textsf{IsCategory} uses a tiered class match (the entity's YOLOE category when it matches, else a caption fallback), and \textsf{HasAttribute} matches the attribute string against the caption and stored attributes through embedding. Landmarks are bound greedily to their top-scoring candidates. 
    \item \emph{Composite scoring.}
    For each target candidate we reduce its predicate scores to a geometric mean $\bar{g}$ (robust to a single near-zero term) and blend it with the target similarity as $s = s_{\mathrm{sem}}\,[\,(1-w) + w\,\bar{g}\,]$ with $w{=}0.5$, where $s_{\mathrm{sem}}$ is additionally scaled by a soft class factor (multiplied by $0.3$ when the candidate's class does not match the parsed target class). 
    The blend lets spatial predicates re-rank semantically plausible candidates -- pulling an unsatisfied candidate down toward half its similarity -- without letting a single noisy predicate veto a strong semantic match; we deliberately neither hard-floor the score nor take a full product. 
    This realizes the soft conjunction over the query formula introduced in~\cref{subsec:problem_formulation}.
    \item \emph{Projected-view reranking.}
    The top-$5$ candidates are reranked by a customized Qwen3.5-9B reranker over \emph{projected views}: for each candidate we render a candidate-centered schematic panel of the local scene geometry, with the candidate marked in red, the competing top-$K$ objects in blue, and the estimated viewing ``front'' direction in green. The numbered panels, together with the parsed predicates and base scores, are presented to the model, which returns a per-candidate  evidence score in $[0,1]$; candidates are reordered by these scores (Qwen-only ordering, i.e.\ a fusion weight of $1$ on the VLM score, with chain-of-thought disabled). This projected-view rerank is our reported configuration (\emph{\modelname}); omitting it gives the \emph{\modelnamebase} ablation.
    \item \emph{Output.}
    The system returns a ranked list of \texttt{ScoredCandidate} records -- a target entity, its composite score, the per-predicate score breakdown, and the bound landmark entities. The action interface $\Psi_t$ converts the top entity into a navigable or viewing pose from the entity's geometry and high-quality views.
\end{enumerate}

\paragraph{Why soft, not hard?}
\label{para:error_decomposition}
Separating candidate generation from predicate scoring lets us reason about two distinct failure modes that our experiments evaluate separately.
A \emph{candidate-recall} failure occurs when the true target is poorly mapped or never enters the candidate set retrieved from its description.
A \emph{relational-ranking} failure occurs when the target is in the set but noisy anchors, predicates, or reranking fail to surface it in the top-$L$.
What keeps these modes independent is that relations enter as soft scores rather than hard filters: with $\operatorname{Agg} = s_{\mathrm{sem}} \cdot \big[ (1 - w) + w\,\bar{g} \big]$, where $s_{\mathrm{sem}} = s_{\sigma_\star}(e)$ is the target similarity, $\bar{g}$ is the geometric mean of predicate scores, and $w = 0.5$, the spatial factor stays in $[1 - w,\, 1]$, so a wrong anchor or noisy predicate can at most halve a candidate's semantic score but never deletes it.
Anchor errors therefore stay out of the recall term and only perturb the ranking term.
In contrast, hard-filtering on all $m$ anchors before scoring causes target recall to degrade multiplicatively with the number of anchors.

\section{\modelname-Scenes}
\label{app:farm-scenes}

\begin{figure}[!h]
    \centering
    \includegraphics[width=1\linewidth]{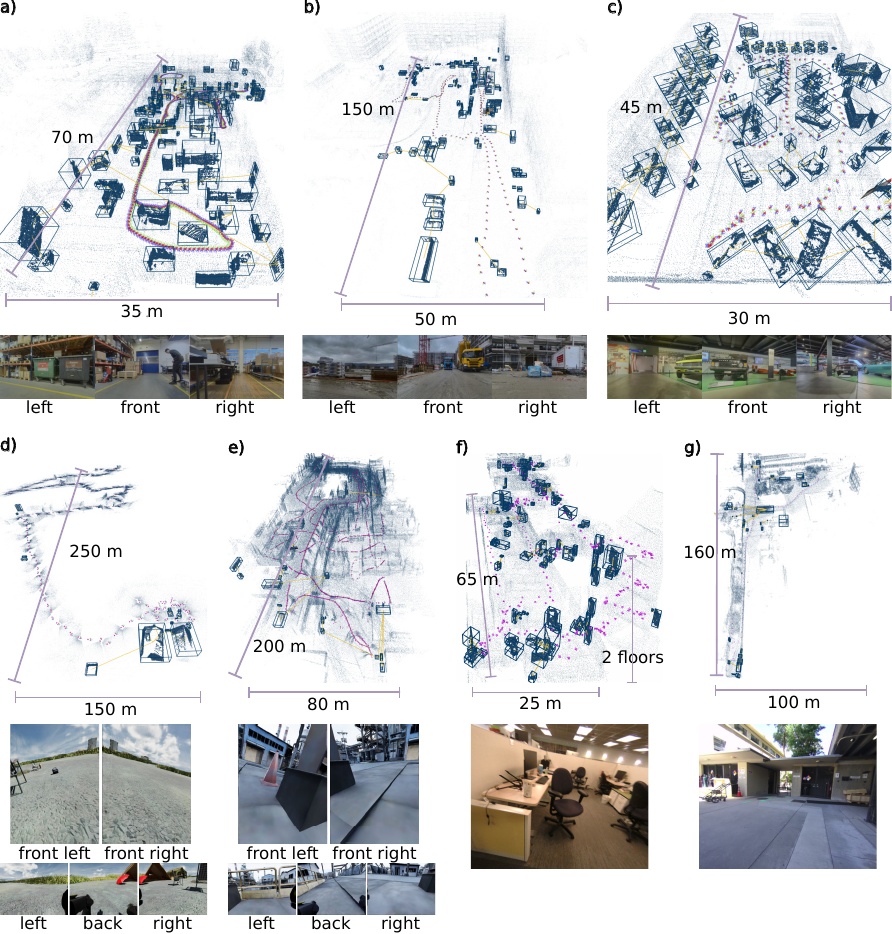}
    \caption{\textbf{\modelname-Scenes.}
    The seven \modelname-Scenes cover diverse indoor and outdoor environments:
    (a) a warehouse~\citep{frey_tuna2026grandtour},
    (b) an outdoor construction site~\citep{frey_tuna2026grandtour},
    (c) an automotive museum~\citep{frey_tuna2026grandtour},
    (d) a camping site,
    (e) an outdoor industrial facility,
    (f) a multi-floor office building, and
    (g) a school campus.
    For each scene, we show the reconstructed point cloud, annotated object instances and relational annotations, robot camera trajectory, approximate scene extent, and representative RGB views. These scenes vary substantially in scale, object density, distractor frequency, and sensor configuration. The large outdoor scenes test online memory construction and retrieval over long trajectories, while the object-rich indoor scenes stress relational retrieval in the presence of many visually or semantically similar distractors.}
    \label{fig:farm-scenes-demo}
\end{figure}

\paragraph{Overview.}
\modelname-Scenes is a curated benchmark of seven large-scale indoor and outdoor environments for evaluating online 3D memory construction and relational object retrieval, as illustrated in~\cref{fig:farm-scenes-demo}. The benchmark spans construction, museum, warehouse, camping, industrial, office, and campus environments, covering a broad range of spatial scales, object densities, camera configurations, and distractor patterns. We summarize the scene statistics in~\cref{tab:farm-scenes-data-card}.

\begin{table*}[t]
\centering
\small
\renewcommand{\arraystretch}{1.08}
\setlength{\tabcolsep}{5pt}
\caption{\textbf{\modelname-Scenes data card.}
For each scene, we report the environment type, setting, approximate area, approximate robot trajectory length, number of cameras, number of RGB-D frames, annotated object instances, and relational language queries.}
\label{tab:farm-scenes-data-card}
\resizebox{\textwidth}{!}{
\begin{tabular}{cllcccccc}
\toprule
\textbf{\#} &
\textbf{Type} &
\textbf{Setting} &
\textbf{Area (m$^2$)} &
\textbf{Trajectory} &
\textbf{Cameras} &
\textbf{Frames} &
\textbf{Annotated Objects} &
\textbf{Queries} \\
\midrule
a) & warehouse                  & Indoor  & 4{,}000  & 205 m   & 3 & 2{,}553 & 163 & 250 \\
b) & construction site          & Outdoor & 10{,}000 & 454 m   & 3 & 1{,}122 & 147 & 145 \\
c) & automotive museum           & Indoor  & 1{,}800  & 160 m   & 3 & 513     & 95  & 203 \\
d) & camping site               & Outdoor & 7{,}500  & 683 m   & 5 & 1{,}274 & 19  & 24  \\
e) & outdoor industrial facility & Outdoor & 15{,}000 & 1{,}267 m & 5 & 2{,}270 & 28  & 50  \\
f) & multi-floor office building & Indoor  & 2{,}400  & 397 m   & 1 & 352     & 147 & 228 \\
g) & school campus              & Outdoor & 6{,}000  & 308 m   & 1 & 210     & 46  & 55  \\
\bottomrule
\end{tabular}
}
\end{table*}

\paragraph{Annotation procedure.}
We annotate objects using an interactive Viser-based interface~\citep{yi2025viser}. For each object of interest, an annotator selects 2D masks in one or more RGB frames, assisted by SAM2~\citep{kirillov2023segany} mask propagation. Masks corresponding to the same physical object across frames are projected into 3D using camera poses and depth, and we extract the object's 3D bounding box from the resulting point cloud. Given the annotated objects, annotators then label relations by selecting object pairs and assigning the most appropriate spatial relation between them. Relations are chosen to be useful for retrieval, so that they can help disambiguate a target object from nearby, visually similar, or semantically similar distractors. We use Gemini~\citep{team2024gemini} to assist with captioning ground-truth objects; all captions are manually verified and refined.

\paragraph{Scene selection rationale.}
We choose the construction site, camping site, outdoor industrial facility, and school campus to evaluate online memory construction and retrieval in large-scale outdoor environments with long trajectories. We choose the automotive museum, warehouse, and multi-floor office building to stress-test relational retrieval in object-rich indoor scenes with many distractors, including repeated cars in the museum, boxes and storage items in the warehouse, and repetitive furniture and equipment in the office building. Together, these scenes test whether a system can maintain an object-level memory online and use relational constraints to retrieve specific objects in cluttered, repetitive, and large-scale environments.

\section{Additional Experiment Results}

\subsection{Datasets}
\label{app:datasets}

We evaluate on two complementary indoor referring-expression benchmarks:
\emph{ReferIt3D}~\note{\cite{achlioptas2020referit_3d}} (NR3D + SR3D+) on ScanNet, and
\emph{IRef-VLA}~\note{\cite{zhang2025iref}} on HM3D.
The ScanNet split emphasizes natural-language robustness: NR3D contains human-written descriptions, while SR3D+ contains synthetic relational descriptions over visually similar distractors.
The HM3D split emphasizes scene scale: our selected HM3D scenes have, on average, $\sim 58\times$ larger volume and $\sim 34\times$ more annotated objects than the selected ScanNet scenes.

To stress representation scalability rather than report aggregate numbers dominated by small rooms, we curate the top 30 scenes from each benchmark ranked by scene complexity: number of annotated objects for ScanNet and number of regions for HM3D.
For each scene, we cap the number of evaluated utterances at 1000, using deterministic stratified sampling to preserve the natural query distribution within each scene.
On ScanNet, we stratify by \texttt{dataset} (NR3D or SR3D+) and \texttt{instance\_type}; on HM3D, we stratify by \texttt{region\_id} and \texttt{relation\_type}.
Scenes with fewer than 1000 utterances contribute all available utterances.
This gives 43{,}076 indoor utterances: 13{,}076 from ScanNet and 30{,}000 from HM3D.

We further introduce \modelname-Scenes, a curated benchmark of seven large-scale indoor and outdoor environments captured with three platforms.
Unlike ScanNet and HM3D, which are reconstructed from indoor houses, \modelname-Scenes includes larger, more cluttered, and more diverse robot-scale deployments with outdoor distractors, long trajectories, and heterogeneous camera configurations.
Together, ScanNet, HM3D, and \modelname-Scenes form a 67-scene benchmark with 44{,}031 utterances, summarized in \cref{tab:dataset_utterances}.

\begin{table}[!h]
\centering
\small
\setlength{\tabcolsep}{6pt}
\caption{
\textbf{Evaluation queries by benchmark split.}
ScanNet and HM3D provide indoor referring expressions from ReferIt3D and IRef-VLA, respectively.
\modelname-Scenes adds large-scale indoor/outdoor queries collected across three capture platforms.
}
\label{tab:dataset_utterances}
\begin{tabular}{llrr}
\toprule
\textbf{Split} & \textbf{Query source} & \textbf{Scenes} & \textbf{Utterances} \\
\midrule
ScanNet           & ReferIt3D (NR3D + SR3D+) & 30 & 13{,}076 \\
HM3D              & IRef-VLA                 & 30 & 30{,}000 \\
\modelname-Scenes & Curated (large-scale)    &  7 &    955 \\
\midrule
\textbf{Total}    &                          & \textbf{67} & \textbf{44{,}031} \\
\bottomrule
\end{tabular}
\end{table}

\paragraph{HM3D RGB-D trajectories.}
ScanNet provides posed RGB-D sequences directly.
HM3D provides textured meshes, so we render a posed RGB-D trajectory for each scene using \texttt{habitat-sim} 0.2.5.
For each annotated ground-truth object, we sample 24 candidate viewpoints within $r\in[0.8,3.0]$\,m of the object centroid on the scene navmesh, select the closest viewpoint without severe occlusion, group accepted viewpoints by navmesh island to cover multi-floor scenes, and order viewpoints within each island by a greedy nearest-unvisited tour.
Consecutive viewpoints are connected by navmesh-constrained shortest paths computed with Habitat-Sim's pathfinder~\cite{puig2023habitat3}, and the resulting polylines are sampled every 0.15\,m to produce dense intermediate camera poses.
Across the 30 HM3D scenes, this procedure gives 99\% coverage of fair ground-truth objects within 2\,m of at least one rendered camera pose, where fair ground-truth objects exclude uninformative categories such as walls and floors.

\subsection{Algorithms}
\label{app:baseline_details}

We compare \modelname against map-based, video-only, and keyframe/agent baselines that represent different design choices for relational object retrieval.
BBQ evaluates an object-centric scene-graph pipeline with LLM-based relational reasoning.
RynnBrain evaluates whether a video VLM can ground the target directly from trajectory frames without an explicit persistent map.
DAAAM + RynnBrain tests whether stronger keyframe selection improves the video-only setting.
DAAAM GPT-5-mini evaluates an agentic scene-understanding pipeline over a 4D graph; because of API cost and runtime, we evaluate it on \modelname-Scenes and on 5-scene subsets of ScanNet and HM3D.

\begin{itemize}[leftmargin=*,topsep=2pt,itemsep=2pt]
\item \textbf{\modelname and \modelnamebase}. 
Our online relational spatial memory (\cref{subsec:memory_construction}) is built from streaming posed RGB-D and queried by the predicate-guarded grounding pipeline in \cref{subsec:spatial_grounding}.
Reconstruction uses an open-vocabulary YOLOE detector, an asynchronous Qwen3.5-9B captioner, and a SigLIP2/Qwen3 text-embedding stack.
We report two variants that share the same memory and candidate retrieval but differ in the final ranking step.
\modelnamebase parses each utterance with Qwen3.5-9B into typed predicates, retrieves candidates by fusing text, caption, and visual embeddings, and ranks them with our soft predicate evaluator (\texttt{unified\_soft\_w50}); it does not invoke a VLM at ranking time.
\modelname adds a projected-view Qwen3.5-9B reranker over the top-5 candidates.
Since this reranker only permutes the top-5 candidates, \modelname and \modelnamebase have identical R@5 and R@10 and differ only in Acc@1 and MRR.
The same grounding configuration is used across ScanNet and HM3D, with no benchmark-specific dispatch.

\item \textbf{BBQ}~\note{\cite{linok2025beyond}}. 
BBQ builds an open-vocabulary object-centric scene graph using MobileSAM masks, DINOv2 cross-view association, and a two-stage LLM grounding procedure for target/anchor selection and relation reasoning.
To remove captioner quality as a confound, we replace BBQ's original captioner with the same Qwen3.5-9B captioner used by \modelname; all other components follow the upstream BBQ pipeline.

\item \textbf{RynnBrain}~\note{\cite{damo2026rynnbrain}}. 
RynnBrain-30B is a 30B-A3B MoE video VLM.
Following its interleaved-frame protocol, the model ingests sub-sampled trajectory frames and directly emits one 3D AABB per utterance, without constructing a persistent map.
To fit within the context length, we downsample trajectories to 2 frames per second.
For mask-IoU evaluation, we convert its 2D box output to a segmentation mask using the same SAM2 mask-conversion code used for all methods.

\item \textbf{DAAAM + RynnBrain}~\citep{gorlo2025DAAAM}. 
We use DAAAM as a frame-selection module.
For each scene, DAAAM solves its set-cover/view-quality program and selects $K=128$ keyframes that cover observed fragments.
These keyframes are then provided to RynnBrain as an image-based memory.

\item \textbf{DAAAM GPT-5-mini}~\citep{gorlo2025DAAAM}. 
We evaluate DAAAM's \texttt{SceneUnderstandingAgent}, an OpenAI Responses-API agent over a 4D graph, using \texttt{gpt-5-mini}~\citep{singh2026openaigpt5card} through OpenRouter with \texttt{max\_iterations}$=5$.
\end{itemize}

All methods share the same captioner endpoint when applicable, the same SAM2 mask-conversion code at evaluation time, the same locked utterance set per benchmark, and the same single-GPU container topology, making latency, memory, and accuracy comparisons directly comparable.

\subsection{Evaluation Metrics}
\label{app:evaluation-metrics}

We report four families of metrics, evaluated consistently across methods.

\begin{itemize}[leftmargin=*,topsep=2pt,itemsep=2pt]
\item \emph{Grounding accuracy.}
For each utterance, a method returns either a ranked list of object candidates or, for RynnBrain-style methods, a single predicted AABB.
For ranked methods, we report $\mathrm{Acc@1}$, $\mathrm{R@5}$, $\mathrm{R@10}$, and $\mathrm{MRR}$ over the top-10 candidates.
$\mathrm{Acc@1}$ measures whether the top-ranked prediction is correct, $\mathrm{R@k}$ measures whether any of the top-$k$ predictions is correct, and $\mathrm{MRR}$ is the mean reciprocal rank of the first correct prediction:
\[
    \mathrm{MRR}
    =
    \frac{1}{N}
    \sum_{i=1}^{N}
    \frac{1}{r_i},
\]
where $r_i$ is the rank of the first correct candidate for query $i$, and the contribution is zero if no correct candidate appears in the top 10.
We evaluate correctness using three overlap criteria:
(i) visible-mask IoU, where the ground-truth object is projected into the selected image plane with occlusion handling;
(ii) 2D box IoU, included for compatibility with box-prediction baselines such as RynnBrain; and
(iii) 3D AABB IoU, included for compatibility with graph baselines such as BBQ.
For each criterion, we report thresholds $\tau\in\{0.10,0.25,0.50\}$.

\item \emph{Representation size.}
For graph-based methods, we report the on-disk size of the saved persistent scene-graph artifact in MiB.
For RynnBrain, which does not persist an explicit map, we instead report the estimated bf16 visual-token-cache size for the keyframes actually ingested by the grounding model, as a representative memory footprint for a no-explicit-map baseline.

\item \emph{Mapping latency.}
For graph-based methods, we report wall-clock time per ingested RGB-D frame, averaged over the full trajectory.
For RynnBrain-style methods, we report latency over the frames effectively consumed by the grounding head.

\item \emph{Query latency.}
We report wall-clock time per utterance, measured end-to-end from the query string to a ranked candidate list, or to a single AABB for RynnBrain-style methods.
\end{itemize}

\subsection{Latency and Accuracy vs. Scene Scale}
\label{app:scaling-analysis}

\begin{figure*}[!h]
    \centering

    \begin{subfigure}[t]{0.95\linewidth}
        \centering
        \includegraphics[width=\linewidth]{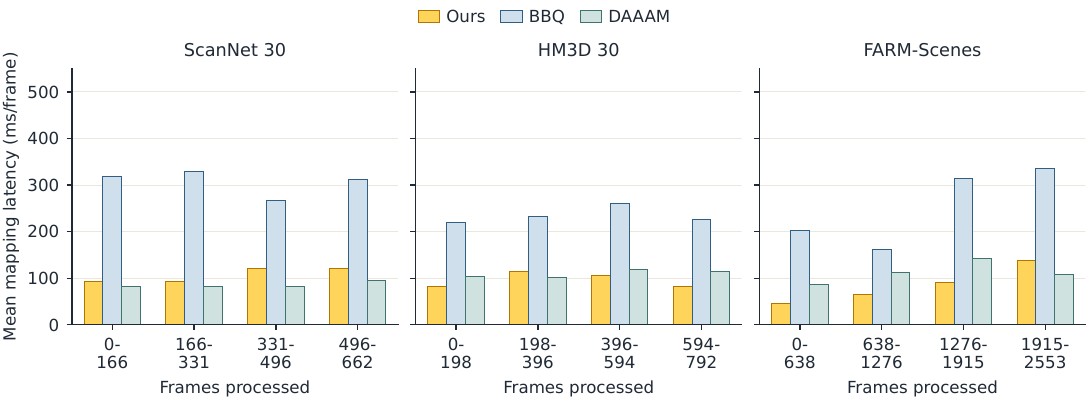}
        \caption{Mapping latency scaling.}
        \label{fig:mapping-latency-scaling}
    \end{subfigure}

    \vspace{0.6em}

    \begin{subfigure}[t]{0.95\linewidth}
        \centering
        \includegraphics[width=\linewidth]{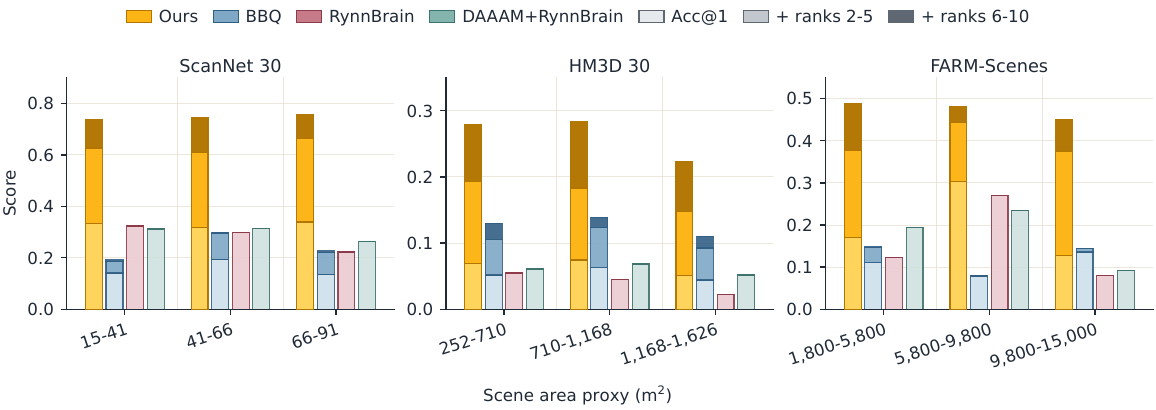}
        \caption{Grounding accuracy versus scene area.}
        \label{fig:accuracy-scene-area}
    \end{subfigure}

    \caption{
    Scaling analysis of online memory construction and grounding performance.
    Top: per-frame mapping latency as a function of distance traveled. 
    Bottom: grounding accuracy as scene area increases.
    }
    \label{fig:scaling-analysis}
\end{figure*}

\cref{fig:scaling-analysis} evaluates whether the online memory construction and grounding behavior remain stable as scene scale increases.
Per-frame mapping latency remains approximately stable as the traversed trajectory length grows (\cref{fig:mapping-latency-scaling}), indicating that the online association and update procedure does not become progressively slower in larger scenes.
Grounding accuracy decreases gradually as scene area increases (\cref{fig:accuracy-scene-area}), which is expected because larger scenes contain more objects, more distractors, and longer-range relational ambiguity.
The degradation is nevertheless smooth rather than catastrophic, supporting the use of \modelname for extended on-device deployments.
Another reason that HM3D scenes have lower accuracy is the lower rendered image quality, which is unrelated to the scene scales.

\subsection{Grounding Accuracy}
\label{app:grounding_accuracy}

\cref{tab:iou_thresholds_combined} reports the main visible-mask IoU evaluation across ScanNet, HM3D, and \modelname-Scenes.
Across all three benchmarks and all IoU thresholds, \modelname achieves the best Acc@1, R@5, R@10, and MRR among methods with ranked outputs.
On ScanNet 30, \modelname reaches Acc@1 of .3591, .2800, and .1304 at IoU thresholds .10, .25, and .50, respectively.
This consistently improves over BBQ and RynnBrain, and the gap becomes especially large at stricter thresholds: at $\tau=.50$, \modelname obtains .1304 Acc@1 compared with .0136 for RynnBrain and .0025 for BBQ.
This suggests that the gain is not only from retrieving roughly relevant objects, but also from selecting candidates whose projected extent better matches the target.

On HM3D 30, the absolute numbers are lower because the scenes are substantially larger, contain more objects and rooms, have lower-quality rendered images, but the same trend holds.
\modelname achieves the best Acc@1 at all thresholds, with .0788, .0606, and .0289 at $\tau=.10,.25,.50$, respectively.
It also substantially improves top-$k$ retrieval: at $\tau=.10$, \modelname reaches .2694 R@10, compared with .1282 for BBQ.
This indicates that the constructed memory often contains the correct target even when final top-1 ranking remains challenging in large multi-room environments.

On \modelname-Scenes, \modelname also performs best across all thresholds.
At $\tau=.10$, it achieves .2419 Acc@1 and .4733 R@10, outperforming BBQ, RynnBrain, DAAAM + RynnBrain, and DAAAM GPT-5-mini.
At $\tau=.25$ and $\tau=.50$, \modelname remains the strongest method, with .1644 and .0754 Acc@1, respectively.
These results are important because \modelname-Scenes contains larger-scale, less curated, and more heterogeneous deployments than the indoor benchmark scenes.

\paragraph{Robustness to evaluation protocol.}
\cref{tab:iou_only_additional_stats} and \cref{tab:3d_aabb_iou_results} evaluate whether the same conclusion holds under alternative overlap definitions.
Under 2D box IoU, \modelname is best on ScanNet at all thresholds and best on HM3D for all top-$k$ and MRR metrics.
The only exception is HM3D Acc@1 at $\tau=.10$, where DAAAM + RynnBrain is slightly higher (.0765 vs. .0705), while \modelname remains better at R@5, R@10, MRR, and at stricter thresholds.
Under 3D AABB IoU, \modelname also outperforms BBQ on both ScanNet and HM3D in mean IoU and in every reported retrieval metric.
These additional protocols show that the improvement is not an artifact of one particular IoU definition.

\paragraph{Natural versus synthetic language.}
\cref{tab:referit3d_split_acc} separates ScanNet performance by ReferIt3D split.
On SR3D+, \modelname obtains the best Acc@1 (.3436), substantially outperforming BBQ (.1628), RynnBrain (.2827), and DAAAM + RynnBrain (.2959).
On NR3D, which contains human-written and more linguistically varied descriptions, \modelname is competitive but slightly below the video-based baselines.
This suggests that the current memory and predicate pipeline is especially effective when the query contains explicit relational structure, while free-form human descriptions remain an important direction for improving semantic parsing and open-vocabulary matching.

Overall, the grounding results support three conclusions.
First, explicit online memory construction combined with symbolic relational grounding yields stronger retrieval than video-only grounding or LLM reasoning over weaker scene graphs.
Second, the advantage persists across indoor datasets, large-scale real-world scenes, and multiple IoU protocols.
Third, the largest gains appear in top-$k$ retrieval and stricter IoU settings, indicating that \modelname improves both candidate recall and spatially precise grounding.

\begin{table*}[!h]
\centering
\scriptsize
\setlength{\tabcolsep}{3.6pt}
\caption{
\textbf{Grounding accuracy under visible-mask IoU.}
We evaluate each method at IoU thresholds $\tau\in\{0.10,0.25,0.50\}$ on ScanNet, HM3D, and \modelname-Scenes.
ScanNet and HM3D use occlusion-aware visible-mask IoU; \modelname-Scenes uses the same 2D mask-IoU protocol on selected evaluation views.
$G$ denotes GPT-5-mini.
}
\label{tab:iou_thresholds_combined}
\begin{tabular}{llrrrrrr}
\toprule
\textbf{Dataset} &
\textbf{Method} &
$n_{\mathrm{scored}}$ &
$\tau$ &
$\mathrm{Acc@1}$ &
$\mathrm{R@5}$ &
$\mathrm{R@10}$ &
$\mathrm{MRR}$ \\
\midrule
\multirow{12}{*}{ScanNet 30}
& \modelname        & 13{,}076 & 0.10 & \textbf{.3591} & \textbf{.6352} & \textbf{.7456} & \textbf{.4794} \\
& RynnBrain-30B     & 13{,}076 & 0.10 & .2797 & -- & -- & -- \\
& BBQ               & 13{,}076 & 0.10 & .1542 & .2284 & .2349 & .1869 \\
& DAAAM + RynnBrain & 11{,}148 & 0.10 & .2929 & -- & -- & -- \\
\cmidrule(lr){2-8}
& \modelname        & 13{,}076 & 0.25 & \textbf{.2800} & \textbf{.5138} & \textbf{.6136} & \textbf{.3825} \\
& RynnBrain-30B     & 13{,}076 & 0.25 & .1395 & -- & -- & -- \\
& BBQ               & 13{,}076 & 0.25 & .0610 & .0915 & .0938 & .0749 \\
& DAAAM + RynnBrain & 11{,}148 & 0.25 & .1323 & -- & -- & -- \\
\cmidrule(lr){2-8}
& \modelname        & 13{,}076 & 0.50 & \textbf{.1304} & \textbf{.2789} & \textbf{.3437} & \textbf{.1973} \\
& RynnBrain-30B     & 13{,}076 & 0.50 & .0136 & -- & -- & -- \\
& BBQ               & 13{,}076 & 0.50 & .0025 & .0041 & .0041 & .0033 \\
& DAAAM + RynnBrain & 11{,}148 & 0.50 & .0125 & -- & -- & -- \\
\midrule
\multirow{12}{*}{HM3D 30}
& \modelname        & 30{,}000 & 0.10 & \textbf{.0788} & \textbf{.1816} & \textbf{.2694} & \textbf{.1254} \\
& RynnBrain-30B     & 30{,}000 & 0.10 & .0458 & -- & -- & -- \\
& BBQ               & 30{,}000 & 0.10 & .0529 & .1073 & .1282 & .0758 \\
& DAAAM + RynnBrain & 26{,}375 & 0.10 & .0610 & -- & -- & -- \\
\cmidrule(lr){2-8}
& \modelname        & 30{,}000 & 0.25 & \textbf{.0606} & \textbf{.1422} & \textbf{.2137} & \textbf{.0978} \\
& RynnBrain-30B     & 30{,}000 & 0.25 & .0325 & -- & -- & -- \\
& BBQ               & 30{,}000 & 0.25 & .0373 & .0789 & .0966 & .0548 \\
& DAAAM + RynnBrain & 26{,}375 & 0.25 & .0443 & -- & -- & -- \\
\cmidrule(lr){2-8}
& \modelname        & 30{,}000 & 0.50 & \textbf{.0289} & \textbf{.0747} & \textbf{.1163} & \textbf{.0500} \\
& RynnBrain-30B     & 30{,}000 & 0.50 & .0138 & -- & -- & -- \\
& BBQ               & 30{,}000 & 0.50 & .0199 & .0419 & .0532 & .0293 \\
& DAAAM + RynnBrain & 26{,}375 & 0.50 & .0188 & -- & -- & -- \\
\midrule
\multirow{15}{*}{FARM-Scenes 7}
& \modelname        & 955 & 0.10 & \textbf{.2419} & \textbf{.3749} & \textbf{.4733} & \textbf{.3046} \\
& RynnBrain-30B     & 955 & 0.10 & .0272 & -- & -- & -- \\
& BBQ               & 955 & 0.10 & .0743 & .0953 & .0963 & .0829 \\
& DAAAM + RynnBrain & 955 & 0.10 & .1508 & -- & -- & -- \\
& DAAAM + G         & 955 & 0.10 & .0560 & .1520 & .2120 & .1000 \\
\cmidrule(lr){2-8}
& \modelname        & 955 & 0.25 & \textbf{.1644} & \textbf{.2681} & \textbf{.3393} & \textbf{.2111} \\
& RynnBrain-30B     & 955 & 0.25 & .0209 & -- & -- & -- \\
& BBQ               & 955 & 0.25 & .0513 & .0649 & .0660 & .0565 \\
& DAAAM + RynnBrain & 955 & 0.25 & .1005 & -- & -- & -- \\
& DAAAM + G         & 955 & 0.25 & .0294 & .0796 & .1090 & .0561 \\
\cmidrule(lr){2-8}
& \modelname        & 955 & 0.50 & \textbf{.0754} & \textbf{.1246} & \textbf{.1665} & \textbf{.0985} \\
& RynnBrain-30B     & 955 & 0.50 & .0094 & -- & -- & -- \\
& BBQ               & 955 & 0.50 & .0293 & .0356 & .0356 & .0320 \\
& DAAAM + RynnBrain & 955 & 0.50 & .0524 & -- & -- & -- \\
& DAAAM + G         & 955 & 0.50 & .0294 & .0754 & .1026 & .0554 \\
\midrule
\multirow{6}{*}{ScanNet 5}
& \modelname & 887 & 0.10 & \textbf{.4059} & \textbf{.6967} & \textbf{.7621} & \textbf{.5310} \\
& DAAAM + G  & 312 & 0.10 & .2630 & .4780 & .5900 & .2710 \\
\cmidrule(lr){2-8}
& \modelname & 887 & 0.25 & \textbf{.3112} & \textbf{.5411} & \textbf{.6212} & \textbf{.4135} \\
& DAAAM + G  & 312 & 0.25 & .2564 & .4679 & .5705 & .3466 \\
\cmidrule(lr){2-8}
& \modelname & 887 & 0.50 & \textbf{.1218} & \textbf{.3416} & \textbf{.3641} & \textbf{.2150} \\
& DAAAM + G  & 312 & 0.50 & .2083 & .4327 & .5417 & .3015 \\
\midrule
\multirow{6}{*}{HM3D 5}
& \modelname & 5{,}000 & 0.10 & \textbf{.0792} & \textbf{.1606} & \textbf{.2424} & \textbf{.1169} \\
& DAAAM + G  & 500 & 0.10 & .0420 & .0940 & .1240 & .0670 \\
\cmidrule(lr){2-8}
& \modelname & 5{,}000 & 0.25 & \textbf{.0586} & \textbf{.1196} & \textbf{.1880} & \textbf{.0876} \\
& DAAAM + G  & 500 & 0.25 & .0400 & .0860 & .1140 & .0622 \\
\cmidrule(lr){2-8}
& \modelname & 5{,}000 & 0.50 & \textbf{.0252} & \textbf{.0590} & \textbf{.0990} & \textbf{.0417} \\
& DAAAM + G  & 500 & 0.50 & .0300 & .0700 & .0900 & .0483 \\
\bottomrule
\end{tabular}
\vspace{-0.5em}
\end{table*}

\begin{table*}[!h]
\centering
\scriptsize
\setlength{\tabcolsep}{4.2pt}
\caption{
\textbf{Grounding accuracy under 2D box IoU.}
We evaluate predictions using $\mathrm{IoU}(\mathrm{pred\_bbox}, \mathrm{GT\_bbox}) \ge \tau$ at thresholds $\tau\in\{0.10,0.25,0.50\}$.
This protocol provides a box-based comparison with methods that directly output bounding boxes.
Results are reported on ScanNet 30 and HM3D 30; DAAAM + RynnBrain is evaluated on the subset of scenes for which its pipeline completed successfully.
}
\label{tab:iou_only_additional_stats}
\begin{tabular}{llrrrrrr}
\toprule
\textbf{Dataset} &
\textbf{Method} &
$n_{\mathrm{scored}}$ &
$\tau$ &
$\mathrm{Acc@1}$ &
$\mathrm{R@5}$ &
$\mathrm{R@10}$ &
$\mathrm{MRR}$ \\
\midrule
\multirow{12}{*}{ScanNet 30}
& \modelname              & 6{,}401 & 0.10 & \textbf{.3376} & \textbf{.6533} & \textbf{.7632} & \textbf{.4706} \\
& RynnBrain-30B     & 8{,}128 & 0.10 & .2239 & -- & -- & -- \\
& BBQ               & 6{,}056 & 0.10 & .1558 & .2248 & .2273 & .1851 \\
& DAAAM + RynnBrain & 7{,}070 & 0.10 & .2348 & -- & -- & -- \\
\cmidrule(lr){2-8}
& \modelname              & 6{,}401 & 0.25 & \textbf{.2371} & \textbf{.4644} & \textbf{.5587} & \textbf{.3343} \\
& RynnBrain-30B     & 8{,}128 & 0.25 & .0335 & -- & -- & -- \\
& BBQ               & 6{,}056 & 0.25 & .0421 & .0595 & .0624 & .0500 \\
& DAAAM + RynnBrain & 7{,}070 & 0.25 & .0344 & -- & -- & -- \\
\cmidrule(lr){2-8}
& \modelname              & 6{,}401 & 0.50 & \textbf{.0767} & \textbf{.1984} & \textbf{.2404} & \textbf{.1263} \\
& RynnBrain-30B     & 8{,}128 & 0.50 & .0019 & -- & -- & -- \\
& BBQ               & 6{,}056 & 0.50 & .0023 & .0041 & .0041 & .0031 \\
& DAAAM + RynnBrain & 7{,}070 & 0.50 & .0023 & -- & -- & -- \\
\midrule
\multirow{12}{*}{HM3D 30}
& \modelname              & 4{,}818 & 0.10 & .0705 & \textbf{.1899} & \textbf{.2824} & \textbf{.1235} \\
& RynnBrain-30B     & 4{,}385 & 0.10 & .0559 & -- & -- & -- \\
& BBQ               & 4{,}126 & 0.10 & .0616 & .1196 & .1414 & .0862 \\
& DAAAM + RynnBrain & 5{,}087 & 0.10 & \textbf{.0765} & -- & -- & -- \\
\cmidrule(lr){2-8}
& \modelname              & 4{,}818 & 0.25 & \textbf{.0550} & \textbf{.1541} & \textbf{.2286} & \textbf{.0988} \\
& RynnBrain-30B     & 4{,}385 & 0.25 & .0421 & -- & -- & -- \\
& BBQ               & 4{,}126 & 0.25 & .0438 & .0885 & .1070 & .0627 \\
& DAAAM + RynnBrain & 5{,}087 & 0.25 & .0543 & -- & -- & -- \\
\cmidrule(lr){2-8}
& \modelname              & 4{,}818 & 0.50 & \textbf{.0273} & \textbf{.0879} & \textbf{.1381} & \textbf{.0545} \\
& RynnBrain-30B     & 4{,}385 & 0.50 & .0163 & -- & -- & -- \\
& BBQ               & 4{,}126 & 0.50 & .0242 & .0511 & .0645 & .0356 \\
& DAAAM + RynnBrain & 5{,}087 & 0.50 & .0208 & -- & -- & -- \\
\bottomrule
\end{tabular}
\vspace{-0.5em}
\end{table*}

\begin{table*}[!h]
\centering
\scriptsize
\setlength{\tabcolsep}{4.2pt}
\caption{
\textbf{Grounding accuracy under 3D AABB IoU.}
We compare graph-based methods using 3D axis-aligned bounding-box overlap at thresholds $\tau\in\{0.10,0.25,0.50\}$.
Mean IoU is computed over top-1 predictions.
We report this metric on ScanNet 30 and HM3D 30; \modelname-Scenes is omitted because sparse and noisy real-world depth makes 3D box IoU unreliable. 
}
\label{tab:3d_aabb_iou_results}
\begin{tabular}{llrrrrrr}
\toprule
\textbf{Dataset} &
\textbf{Method} &
\textbf{Mean IoU} &
$\tau$ &
$\mathrm{Acc@1}$ &
$\mathrm{R@5}$ &
$\mathrm{R@10}$ &
$\mathrm{MRR}$ \\
\midrule
\multirow{6}{*}{ScanNet 30}
& \modelname & \textbf{.1128} & 0.10 & \textbf{.2944} & \textbf{.5995} & \textbf{.7061} & \textbf{.4275} \\
& BBQ  & .0969          & 0.10 & .2271          & .3536          & .3585          & .2807 \\
\cmidrule(lr){2-8}
& \modelname & \textbf{.1128} & 0.25 & \textbf{.2010} & \textbf{.4686} & \textbf{.5743} & \textbf{.3216} \\
& BBQ  & .0969          & 0.25 & .1578          & .2502          & .2589          & .1978 \\
\cmidrule(lr){2-8}
& \modelname & \textbf{.1128} & 0.50 & \textbf{.0779} & \textbf{.2112} & \textbf{.2668} & \textbf{.1360} \\
& BBQ  & .0969          & 0.50 & .0769          & .1306          & .1367          & .1007 \\
\midrule
\multirow{6}{*}{HM3D 30}
& \modelname & \textbf{.0148} & 0.10 & \textbf{.0402} & \textbf{.1183} & \textbf{.1793} & \textbf{.0849} \\
& BBQ  & .0129          & 0.10 & .0305          & .0648          & .0794          & .0456 \\
\cmidrule(lr){2-8}
& \modelname & \textbf{.0148} & 0.25 & \textbf{.0233} & \textbf{.0731} & \textbf{.1140} & \textbf{.0529} \\
& BBQ  & .0129          & 0.25 & .0193          & .0412          & .0524          & .0290 \\
\cmidrule(lr){2-8}
& \modelname & \textbf{.0148} & 0.50 & \textbf{.0094} & \textbf{.0296} & \textbf{.0452} & \textbf{.0216} \\
& BBQ  & .0129          & 0.50 & .0091          & .0205          & .0279          & .0143 \\
\bottomrule
\end{tabular}
\vspace{-0.5em}
\end{table*}

\begin{table}[!h]
\centering
\small
\setlength{\tabcolsep}{5pt}
\caption{
\textbf{ScanNet grounding accuracy by ReferIt3D split.}
We report Acc@1 under visible-mask IoU at $\tau=0.10$ separately on NR3D, SR3D+, and their combined ScanNet split.
NR3D contains human-written referring expressions, while SR3D+ contains synthetic relational descriptions.
}
\label{tab:referit3d_split_acc}
\begin{tabular}{lrrrrrr}
\toprule
\textbf{Method} &
\textbf{NR3D} & \textbf{$n$} &
\textbf{SR3D+} & \textbf{$n$} &
\textbf{Combined} & \textbf{$n$} \\
\midrule
\modelname &
0.2520 & 1{,}893 &
\textbf{0.3436} & 11{,}183 &
\textbf{0.3304} & 13{,}076 \\
BBQ &
0.1030 & 1{,}893 &
0.1628 & 11{,}183 &
0.1542 & 13{,}076 \\
RynnBrain-30B &
0.2620 & 1{,}893 &
0.2827 & 11{,}183 &
0.2797 & 13{,}076 \\
DAAAM + RynnBrain &
\textbf{0.2712} & 1{,}357 &
0.2959 & 9{,}791 &
0.2929 & 11{,}148 \\
\bottomrule
\end{tabular}
\vspace{-0.5em}
\end{table}

\subsection{Failure Cases}
\label{app:failure-analysis}

\begin{figure}[!h]
    \centering
    \includegraphics[width=1.0\linewidth]{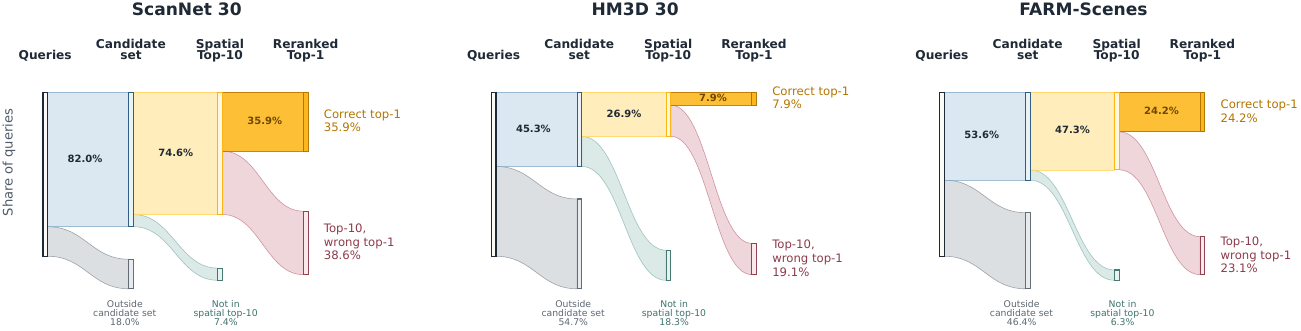}
    \caption{
    \textbf{Where grounding failures occur.}
    We decompose each query into three stages: whether the ground-truth object appears in the retrieved candidate set, whether it survives spatial reasoning into the top-10 list, and whether the final reranker places it at rank 1.
    Most ScanNet failures occur at the final reranking stage, while HM3D and \modelname-Scenes failures are dominated by missing targets from the initial candidate set.
    }
    \label{fig:failure-location-sankey}
\end{figure}

\cref{fig:failure-location-sankey} localizes failures along the retrieval pipeline.
On ScanNet, the ground-truth object appears in the candidate set for 82.0\% of queries and remains in the spatial top-10 for 74.6\%, but only 35.9\% are ranked first after reranking.
Thus, the main remaining bottleneck on ScanNet is not memory recall, but fine-grained top-1 disambiguation among plausible candidates.
In particular, 38.6\% of all queries contain the correct object in the spatial top-10 but are assigned the wrong top-1 prediction, suggesting that improved view selection, visual scoring, or query-conditioned reranking could further improve performance.

HM3D exhibits a different failure mode.
Only 45.3\% of queries contain the ground-truth object in the candidate set, and 54.7\% fail before spatial reasoning can be applied.
This indicates that the primary bottleneck on HM3D is memory construction and candidate recall rather than relational reasoning alone.
HM3D is also more challenging because RGB-D trajectories are rendered from reconstructed meshes, whose image quality, texture fidelity, and object visibility are generally lower than real RGB-D scans.
This can degrade open-vocabulary detection, captioning, and visual embedding quality, especially in large multi-room scenes with many small or partially visible objects.

\modelname-Scenes lies between these two regimes.
The correct object appears in the candidate set for 53.6\% of queries and survives into the spatial top-10 for 47.3\%, showing that spatial filtering removes relatively few additional targets once they are retrieved.
However, 23.1\% of queries still contain the correct object in the top-10 but fail at final top-1 selection.
This suggests two complementary directions for improvement: stronger online object discovery to reduce candidate-set misses in large real-world scenes, and more reliable final reranking to distinguish the target from visually or semantically similar distractors.

\subsection{Real Robot Experiment}
\label{app:real-robot}

To validate that \modelname can build and query its memory \emph{online and on-device} during a physical deployment, we run a real-robot trial in which a quadruped first explores an environment under teleoperation and then autonomously retrieves and navigates to objects specified by free-form relational language. Crucially, the memory is constructed on the robot's onboard compute through the same streaming pipeline used in live operation, and queries are issued only after mapping is complete, so the experiment exercises online construction, relational retrieval, and object-goal navigation end-to-end under real-world sensing and timing.

\begin{figure}
    \centering
    \includegraphics[width=1.0\linewidth]{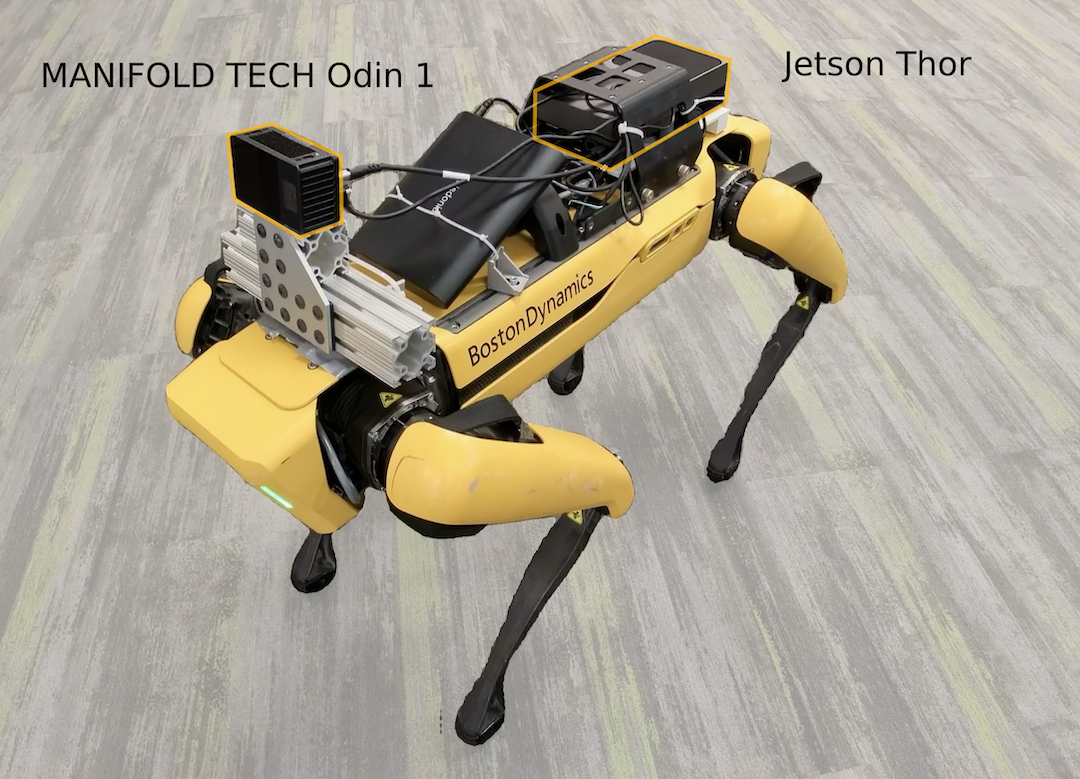}
    \caption{\textbf{Real-robot platform.} Boston Dynamics Spot with a Manifold Tech Odin~1 perception unit (RGB-D and LiDAR visual-inertial odometry) and an onboard NVIDIA Jetson Thor running \modelname.}
    \label{fig:real-robot-setup}
\end{figure}

\parsection{Robot platform and sensing}
We use a Boston Dynamics Spot quadruped, equipped with a Manifold Tech Odin~1 perception unit and an NVIDIA Jetson Thor onboard computer (\cref{fig:real-robot-setup}). The Odin~1 is a self-contained sensor that provides RGB, metric depth, and a built-in LiDAR visual-inertial odometry pose stream, removing the need for an external SLAM stack at mapping time. An adapter node converts the raw topics into posed RGB-D frames for \modelname to ingest into the online memory-construction node.

\parsection{Onboard compute and deployment}
All perception and memory construction run on the Jetson Thor, an aarch64 platform with a Blackwell-class GPU and 128\,GB of unified CPU--GPU memory: the segmentation and embedding models run in-process, and the Qwen3.5-9B captioner and Qwen3 embedding model are served by a local vLLM server on the same device~\citep{qwen35blog, qwen3vlembedding, team2026qwen3, kwon2023efficient}. We build on an existing ROS~2 base stack that provides the Spot driver and the Odin~1 sensor driver; \modelname runs as a separate process alongside it on a shared DDS domain, communicating only through ROS~2 topics, so the two stacks need not be merged. The mapping node always processes the newest available frame and drops any stale frames that arrive while it is busy, enabling it to keep up with real-time operation.

\begin{figure}
    \centering
    \includegraphics[width=1.0\linewidth]{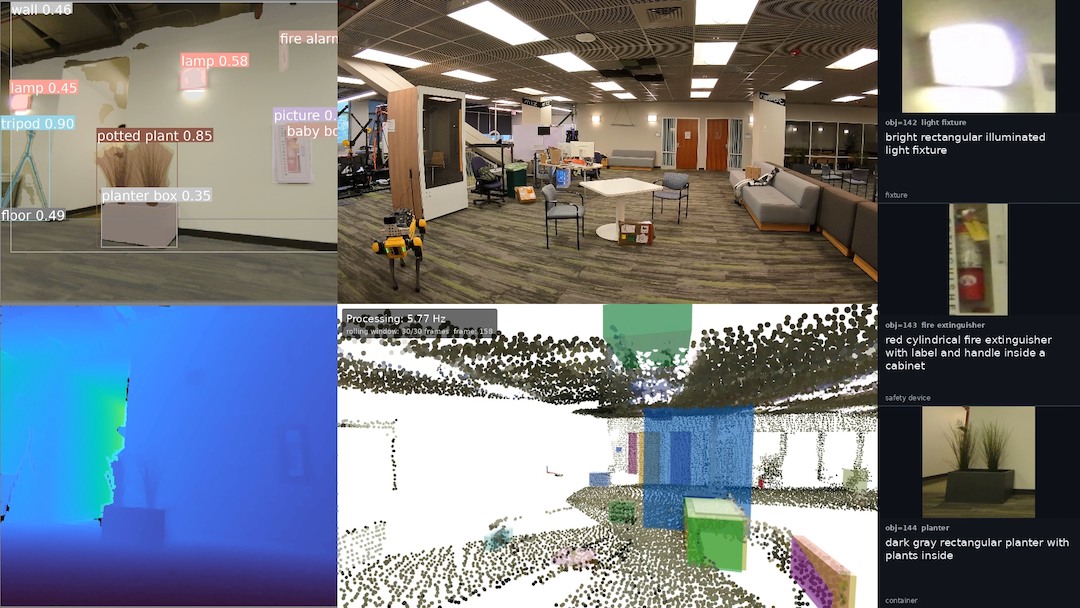}
    \caption{\textbf{Onboard scene-graph construction.} A snapshot of \modelname building its object-level memory on the robot from the Stage-1 log. \emph{Left}: an RGB frame with open-vocabulary detections and its aligned metric depth. \emph{Middle}: the mapped environment and the accumulated 3D point clouds with per-object bounding boxes. \emph{Right}: example object-memory entries, each with a category, a crop, and the VLM caption used for retrieval.}
    \label{fig:real-robot-construction}
\end{figure}

\parsection{Stage 1: memory construction from a teleoperation log}
In the first stage, an operator manually teleoperates Spot and the onboard Odin~1 unit through the environment, recording the raw sensor streams to a rosbag. The bag is later replayed on the Jetson Thor through the same online mapping node used for live operation, which incrementally constructs the object-level memory of~\cref{app:memory-construction} and persists it as a scene state. Replaying through the online node rather than an offline driver reproduces a live run's behavior, while recording lets us reuse a single capture for reproducibility. The saved scene state also records the mapping-time camera poses, which Stage~2 reuses to construct a simple navigation graph.

\parsection{Stage 2: language-conditioned object-goal navigation}
In the second stage the robot operates autonomously over the constructed scene memory, given spatial-relational object-query strings. Each query is grounded against the onboard memory by the retrieval pipeline of~\cref{app:grounding-details}, returning the top-ranked object and a saved viewpoint from which it was observed during mapping. Everything runs on the robot's Jetson Thor, including the onboard vLLM server that hosts the language models used for parsing and embedding.

Because the experiment is meant to highlight the contributions of online mapping and relational retrieval, we pair them with a deliberately simple, graph-based navigation stack rather than a general-purpose planner. We retrace the Stage-1 trajectory, treating its recorded waypoints as a navigation graph: given a retrieved target we take its saved viewpoint as the goal, plan a shortest path via Dijkstra, and execute it by streaming waypoints to Spot's trajectory controller.

\parsection{Qualitative demonstration}
Rather than a large-scale quantitative benchmark, this trial serves as a qualitative demonstration that \modelname's onboard memory supports relational selection and that Spot can act on the result by navigating to the retrieved object's saved viewpoint. We issued the small set of relational queries listed in~\cref{tab:placeholder-real-robot}, each exercising a different spatial predicate. Several queries deliberately share a target category -- in particular multiple cardboard packages -- so the correct referent can only be singled out by the relational constraint rather than by category alone. For each query, \modelname parses the utterance, grounds the target against the memory built in Stage~1, and returns its saved viewpoint, after which the robot navigates to that viewpoint; \cref{fig:real-robot-construction} shows representative episodes. \note{Summarize the observed outcomes qualitatively -- which queries selected the intended object and drove Spot to the right viewpoint, plus any notable failure and its cause (candidate recall, relational ranking, or odometry drift), tying back to~\cref{para:error_decomposition}.} More details of the real robot experiments are in the supplementary video.

\begin{table}[!h]
\centering
\small
\setlength{\tabcolsep}{6pt}
\caption{\textbf{Real-robot demonstration queries.}
The relational utterances used in the on-device trial, each exercising a different spatial predicate. Several share a target category (e.g., cardboard package), so the referent can be disambiguated only by the relation. For each, \modelname grounds the target in the onboard memory and Spot navigates to its saved viewpoint.}
\label{tab:placeholder-real-robot}
\begin{tabular}{llc}
\toprule
\textbf{Query} & \textbf{Relation} & \textbf{Outcome} \\
\midrule
``door to the right of the sofa''                      & \textsf{RightOf}   & \checkmark \\
``cardboard package in front of the trash can''         & \textsf{InFrontOf} & \checkmark \\
``cardboard package under a table''               & \textsf{Below}     & \xmark \\
``cardboard package under a white table''               & \textsf{Below}     & \checkmark \\
``cardboard package next to a humanoid robot''          & \textsf{NextTo}    & \checkmark \\
``chair between a white table and a robot'' & \textsf{Between}   & \xmark \\
``gray fabric chair between a white table and a robot'' & \textsf{Between}   & \checkmark \\
``purple office chair closest to a package''            & \textsf{Closest}   & \checkmark \\
\bottomrule
\end{tabular}
\end{table}








\end{document}